\documentclass[journal]{IEEEtran}
\usepackage{amsmath}
\usepackage{amsfonts}
\usepackage{graphicx}
\usepackage[utf8]{inputenc}
\usepackage{amsthm}
\usepackage{amssymb}
\usepackage{array}
\usepackage{notoccite}
\usepackage{booktabs}
\usepackage{algpseudocode}
\usepackage{color}
\usepackage{array}
\usepackage{float}
\usepackage{tabularray}
\UseTblrLibrary{booktabs}
\usepackage{xcolor}
\usepackage{siunitx}
\usepackage{multirow}
\usepackage{subfig}
\usepackage{graphicx}
\usepackage[ruled,vlined]{algorithm2e}
\usepackage{graphicx}
\usepackage{amsthm}
\usepackage{hyperref}
\usepackage{comment}
\usepackage{tabularx}
\usepackage{multirow}
\usepackage{float}

\theoremstyle{definition}

\theoremstyle{remark}

\begin{document}

\title{Autonomous Tracking and Terminal Guidance of Moving Targets for Fixed-Wing UAVs\thanks{Department of Mechanical Engineering, National Yang Ming Chiao Tung
University, Hsinchu, Taiwan 30010 Email: howard55.en12@nycu.edu.tw and tenghu@nycu.edu.tw}\thanks{This research was supported by the Ministry of Science and Technology,
Taiwan (Grant Number NSTC 114-2223-E-A49 -004 -MY3}}

\author{Wei-Hao Liou and Teng-Hu Cheng}

\maketitle 

\begin{abstract}

This study introduces a unified control framework for fixed-wing unmanned aerial vehicles (UAVs) fitted with a pan-tilt (PT) camera, intended to perform an end-to-end mission spanning from initial target detection to accurate terminal engagement. The proposed system employs a three-phase strategy: a vision-based target acquisition phase, an NMPC-based tracking phase, and a terminal guidance phase. During tracking, the framework uses an Unscented Kalman Filter (UKF) to fuse YOLO-based visual detections with inertial measurements, enabling robust target state estimation under unknown dynamics. To ensure reliable visual contact, we introduce a constraint-aware Nonlinear Model Predictive Control (NMPC) strategy that incorporates Control Barrier Functions (CBFs) to explicitly prevent UAV self-occlusion---a common limitation in fixed-wing tracking. Upon satisfying terminal engagement conditions, the system seamlessly transitions control to a quaternion-based Biased Proportional Navigation Guidance (BPNG) law, enforcing precise impact angle constraints. High-fidelity simulations demonstrate that the framework achieves stable, robust tracking and accurate terminal interception while strictly respecting the vehicle's dynamic limits and camera field-of-view constraints.

\end{abstract}


\IEEEpeerreviewmaketitle

\section{Introduction}

\subsection{Background}
Unmanned aerial vehicles (UAVs) are indispensable in modern surveillance \cite{6761569}, mapping \cite{rs9111187}, search and rescue \cite{1310000, 6290694}, and military operations \cite{CHAMOLA2021102324} due to their high efficiency and cost-effectiveness. Among these, fixed-wing UAVs are preferred for high-speed, long-endurance, and large-area missions. Recent advancements in onboard vision have expanded their operational scope to include real-time target tracking. However, tracking moving targets with fixed-wing platforms is inherently challenging due to coupled roll-yaw dynamics and restricted lateral maneuverability. While pan-tilt (PT) cameras improve sensing flexibility, they introduce complex gimbal dynamics and physical field-of-view (FOV) constraints that must be accounted for in the control design. Furthermore, precision missions often require terminal guidance—such as achieving a specific impact angle—which poses significant difficulty for fixed-wing airframes. Our work addresses these requirements by integrating mid-course NMPC-based tracking with terminal BPNG guidance, ensuring both visual contact and precise engagement geometry.

\begin{figure}[H]
    \centering
    \includegraphics[width=\columnwidth]{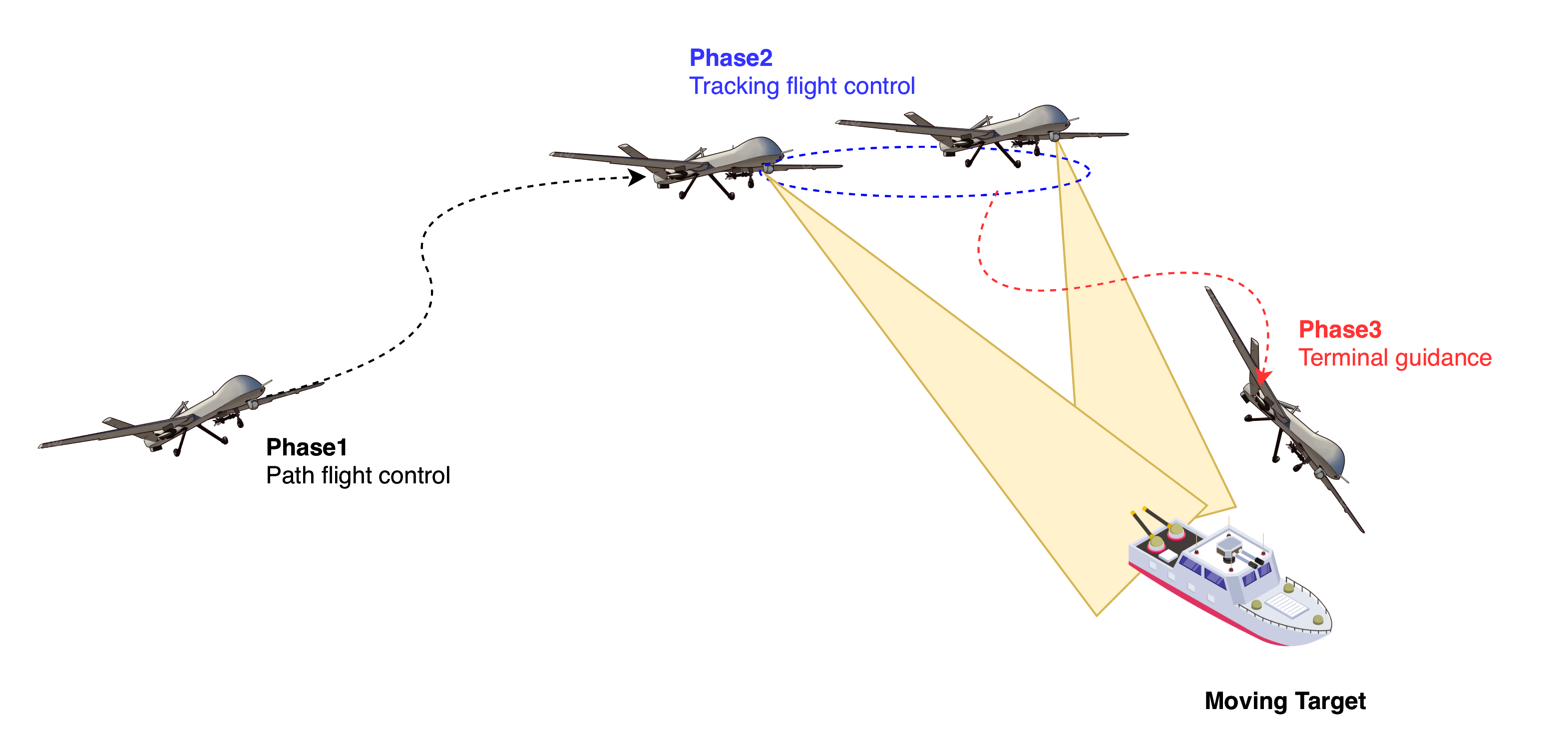}
    \caption{Tracking and terminal mission of a fixed-wing UAV equipped with a PT camera.}
    \label{fig:phases}
\end{figure}

\subsection{Related Works}
Target tracking has been extensively researched, particularly for rotor-type UAVs that benefit from hovering capabilities \cite{Ye2020, GomezBalderas2013, 8407314}. For fixed-wing platforms, tracking is complicated by the lack of hovering, requiring controllers that explicitly model motion constraints \cite{oh2013rendezvous, 5611328}. To increase sensing versatility, researchers have integrated visual feedback \cite{QUINTERO201428} and gimbal cameras \cite{peliti2012vision, yang2020image}. While these studies make significant progress, they often rely on ideal flight assumptions or neglect practical constraints such as feasible speed/altitude envelopes and self-occlusion.

Nonlinear Model Predictive Control (NMPC) is a standard for handling fixed-wing dynamics and constraints \cite{giuffrida2018model, Ulker2017, stastny2017nonlinear, mali2020model, tyagi2021nmpc}. However, many NMPC formulations assume full prior knowledge of the target state, which is often unavailable in real-world scenarios \cite{tyagi2021nmpc}. For terminal engagement, Proportional Navigation Guidance (PNG) and its biased variants (BPNG) are widely used for their simplicity \cite{yang1997unified, adler1956missile, smith2008proportional}. While recent studies have improved BPNG performance through optimal control \cite{kim2021quaternion}, achieving robust terminal guidance often requires extensive gain tuning. Finally, although various estimation techniques—such as particle filters \cite{dong2017maneuvering} and Unscented Kalman Filters (UKF) \cite{wang2014vision}—have been proposed to solve the unknown-target-motion problem, an integrated framework that jointly optimizes fixed-wing dynamics, camera PT constraints, and terminal guidance remains a significant open challenge in the literature.
\subsection{Objective}

The objective of this work is to develop an integrated guidance, control, and estimation framework for fixed-wing UAVs that enables robust visual target tracking and precise terminal guidance under realistic dynamic and sensing constraints. As illustrated in Fig.~\ref{fig:phases}, the mission is divided into three sequential phases: (1) initial path flight, where the UAV transitions from its starting location; (2) tracking control, where the UAV maintains a favorable relative position to the moving target while respecting its dynamic constraints and PT joint limits; and (3) terminal guidance, where the UAV aligns its trajectory to achieve the desired impact while maintaining visual contact with the target. The control objective is to ensure robust tracking and accurate terminal engagement throughout all phases.

\subsection{Contribution}
This study presents an integrated, three-phase autonomous framework for fixed-wing UAVs equipped with a  PT camera to achieve robust, long-duration tracking and precise terminal engagement of unknown moving targets. Unlike existing works that often rely on unrealistic assumptions—such as known target dynamics \cite{tyagi2021nmpc}, ideal flight conditions \cite{yang2020image}, or isolated tracking and guidance loops—our framework explicitly couples nonlinear estimation with constraint-aware path planning and terminal guidance laws.

The primary contributions of this research are as follows:

\begin{enumerate}
    \item \textbf{Unified Estimation and Tracking Framework:} We propose an integrated system that relies solely on visual information to perform real-time estimation and tracking. By incorporating a UKF to fuse YOLO-detected image features with inertial measurements, the system robustly estimates the target's position and velocity under unknown motion, bridging the gap between perception and control.
    
    \item \textbf{Constraint-Aware NMPC for Gimbaled Tracking:} We propose an NMPC framework that ensures target tracking is preserved while simultaneously adhering to the UAV’s dynamic flight envelope and the physical operating constraints of the PT camera. By employing control barrier functions (CBFs), we formulate an explicit constraint to avoid UAV self-occlusion, ensuring continuous visual contact even during aggressive maneuvers—a critical limitation in SOTA tracking approaches that often ignore airframe-shadowing during turns.
    
    \item \textbf{Seamless Transition to Optimal Terminal Guidance:} Our framework features a novel phase-switching logic that transitions from NMPC-based loitering to a quaternion-based biased proportional navigation guidance (BPNG) law. This transition ensures that once mission conditions are met, the UAV is steered to satisfy strict terminal impact angle constraints (IAC) without requiring manual retuning or complex gain scheduling, providing a significant improvement in engagement reliability over classical PNG-based methods.
\end{enumerate}
\section{Problem Formulation}
\subsection{Kinematics of the overall system}
\begin{figure}[h]
    \centering
    \includegraphics[width=\columnwidth]{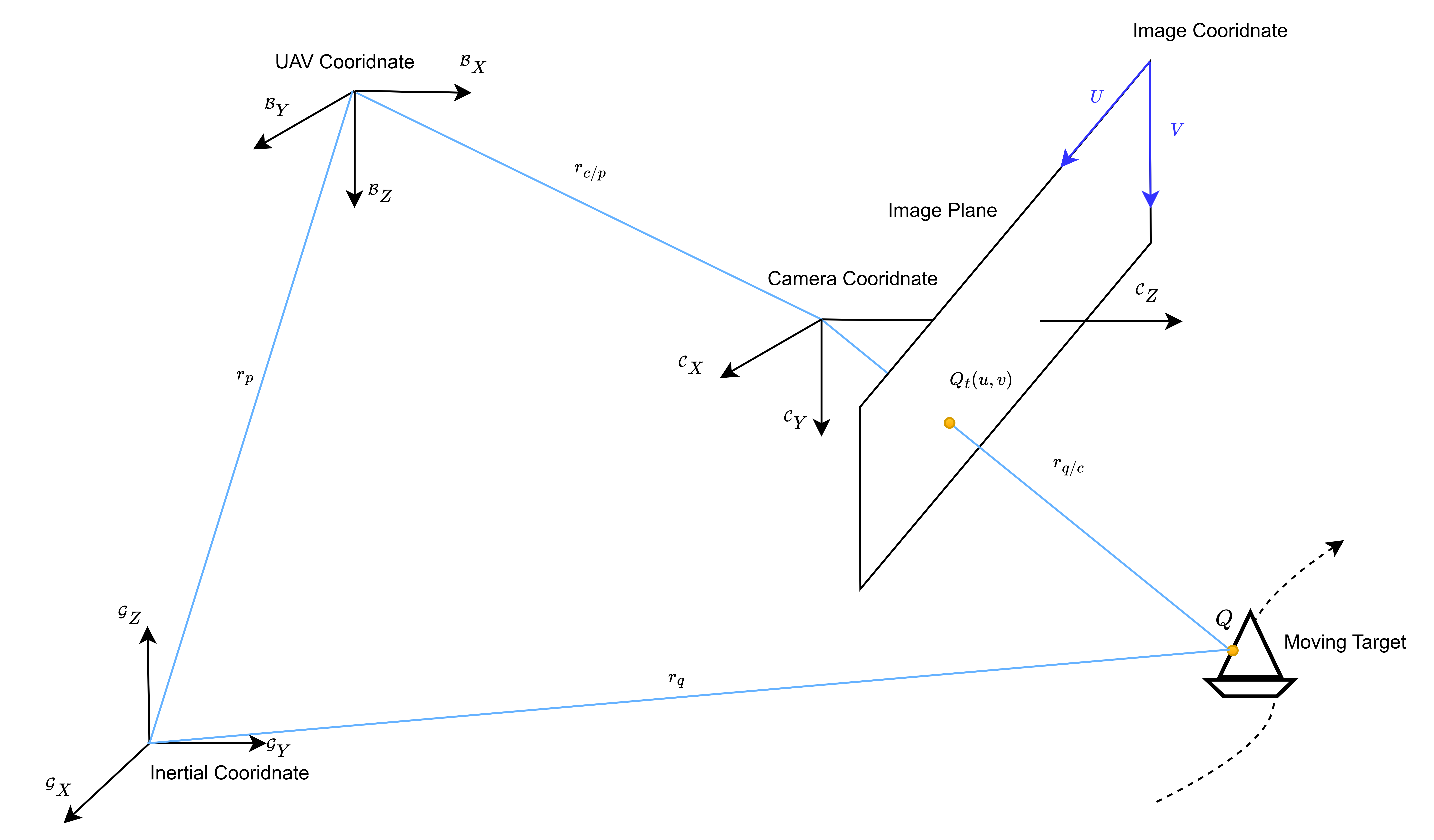}
    \caption{Kinematics model}
    \label{fig:framework}
\end{figure}

The kinematic configuration of the complete system is shown in Fig.~\ref{fig:framework}. The system is composed of a fixed-wing UAV, an onboard PT camera, and the target. Multiple coordinate frames are involved and summarized in Table~\ref{tab:symbol_table}.

\begin{table}[h]
    \begin{center}
        \caption{Definition of Symbol.}
        \begin{tabular}{c c}
        \hline
        Symbol&Description  \\
        \hline
        ${}^\mathcal{G}(\cdot)$&Inertial (ground) frame\\
        $^\mathcal{B}(\cdot)$&UAV body frame\\
        $^\mathcal{C}(\cdot)$&Camera frame\\
        $r\in \mathbb{R}^3$&Position\\
        $\omega\in\mathbb{R}^3$&Angular Velocity\\
        \hline
        \end{tabular}
        \label{tab:symbol_table}
    \end{center}
\end{table}

 The inertial frame, denoted as the ground frame $\mathcal{G}$, follows an East–North–Up (ENU) convention. The UAV body frame $\mathcal{B}$ is defined using a front–right–down (FRD) convention, and under coordinated flight with negligible sideslip, the UAV motion is primarily aligned with the $^\mathcal{B}X$ axis. The camera frame $\mathcal{C}$ is defined with the $^\mathcal{C}Z$ axis pointing forward, $^\mathcal{C}X$ to the right, and $^\mathcal{C}Y$ downward.
\begin{figure}[h]
    \centering
    \includegraphics[width=\columnwidth]{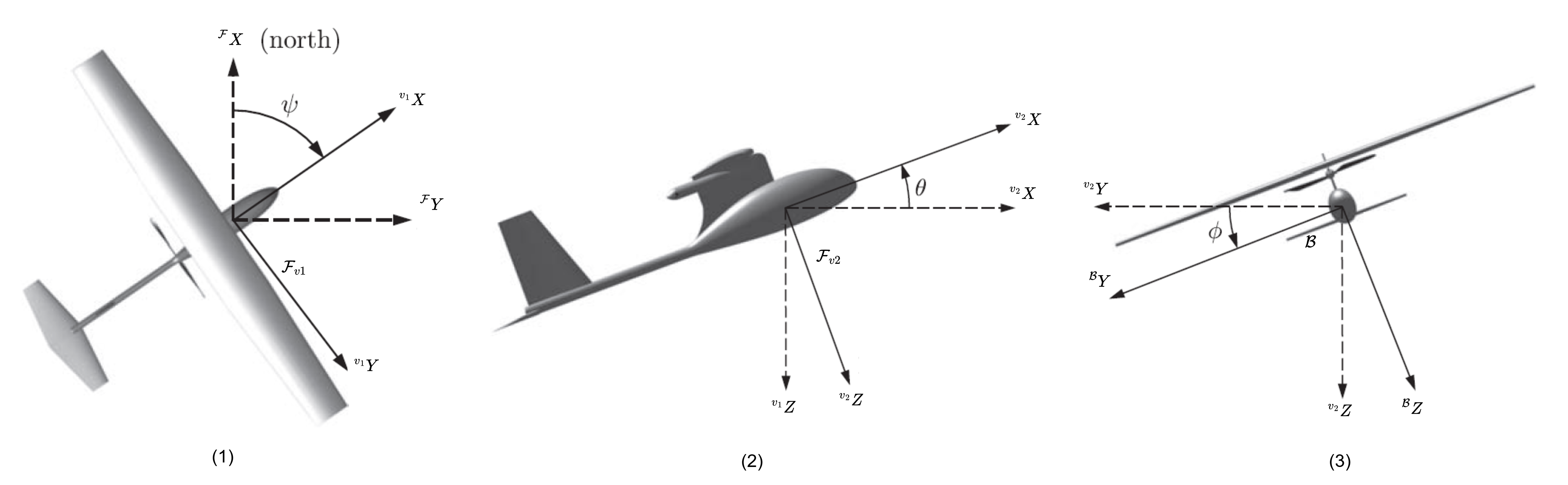}
    \caption{Euler angle definition and rotation order \cite{valencia2012small}.}
    \label{fig:rotation}
\end{figure}
Consider the UAV's Euler angles: roll $\phi$, pitch $\theta$, and yaw $\psi$, as defined in  \cite{valencia2012small}, shown in Fig.~\ref{fig:rotation}. The transformation from the inertial frame to the UAV body frame is given by:
\begin{equation}
    ^\mathcal{B}R_\mathcal{G}=R_x(\phi)R_y(\theta)R_z(\psi){^{NED}R}_{ENU},
    \label{eq:1}
\end{equation}
where ${^{NED}R}_{ENU}$ is a $3\times3$ transformation matrix that converts coordinates from the ENU frame to the North-East-Down (NED) frame. The matrices $R_x$, $R_y$, and $R_z$ denote $3\times3$ rotation matrices corresponding to rotations about the $x$, $y$, and $z$ axes, respectively.
Thus, the UAV velocity in inertial frame $^\mathcal{G}V_p$ is defined as:
\begin{equation}
    ^\mathcal{G}V_p=\begin{bmatrix}^\mathcal{G}v_{px}\\^\mathcal{G}v_{py}\\^\mathcal{G}v_{pz}\end{bmatrix}=\begin{bmatrix}V_U\cos{\psi}\cos{\theta} \\
V_U\sin{\psi}\cos{\theta}\\
V_U\sin{\theta}\end{bmatrix},
\label{eq:2}
\end{equation}
where $V_U$ represents the controlled UAV velocity along its body frame $x$-axis.
In addition, the relative pose from the camera to the target illustrated in Fig. \ref{fig:framework} is given by:
\begin{equation}
    r_{q/c}=r_q-r_p-r_{c/p},
    \label{eq:3}
\end{equation}
where $r_{c/p}$ is the relative position from the UAV coordinate to the PT camera, as shown in Fig. \ref{fig:Gimbalcoord}.
Taking time derivative of (\ref{eq:3}) yields:
\begin{equation*}
    \dot{r}_{q/c}=V_q-V_p-\omega_p\times r_{q/c}-\omega_c\times r_{q/c}-\omega_p\times r_{c/p},
    \label{eq:4}
\end{equation*}
where $V_p=[v_{px}\ v_{py}\ v_{pz}]^T$ is the translational velocity of the UAV, $\omega_p=[\omega_{px}\ \omega_{py}\ \omega_{pz}]^T$ is the angular velocity of the UAV, $\omega_c=[\omega_{cx}\ \omega_{cy}\ \omega_{cz}]^T$ is the camera's angular velocity and $V_q=[v_{qx}\ v_{qy}\ v_{qz}]$ is the target's translational velocity, which is unknown and will be estimated using UKF method.

\subsection{PT Camera Model}
\begin{figure}[h]
    \centering
    \includegraphics[width=\columnwidth]{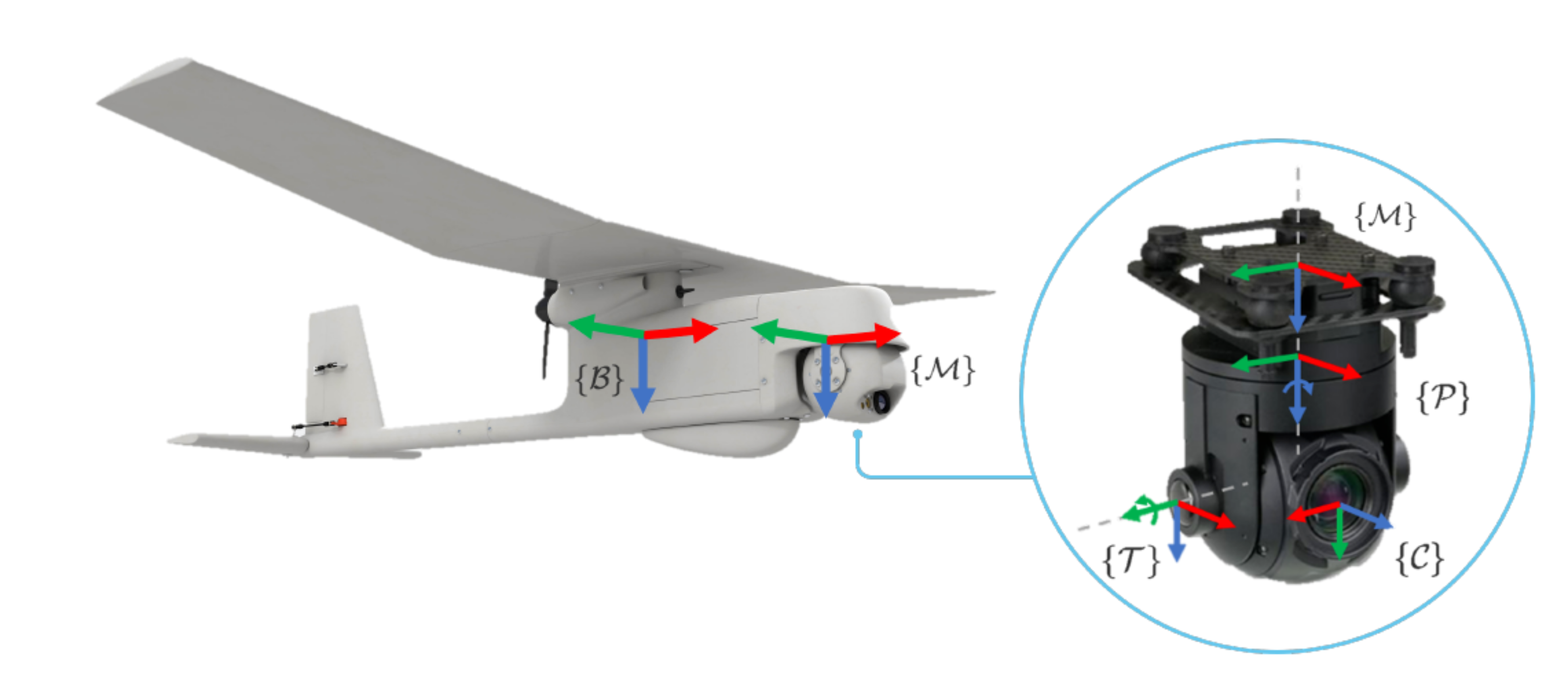}
    \caption{Coordinate systems of a fixed-wing UAV mounted with PT camera.}
    \label{fig:Gimbalcoord}
\end{figure}

To further illustrate the  tracking task and the target estimation derived from camera feedback, a model of the PT camera motion is introduced in this section. The coordinate relationship between the fixed-wing UAV and the PT camera is depicted in Fig. \ref{fig:Gimbalcoord}, where $\mathcal{M}$ denotes the mount frame, $\mathcal{P}$ the pan frame, $\mathcal{T}$ the tilt frame, and $\mathcal{C}$ the camera frame.

\begin{figure}[h]
    \centering
    \includegraphics[width=0.5\columnwidth]{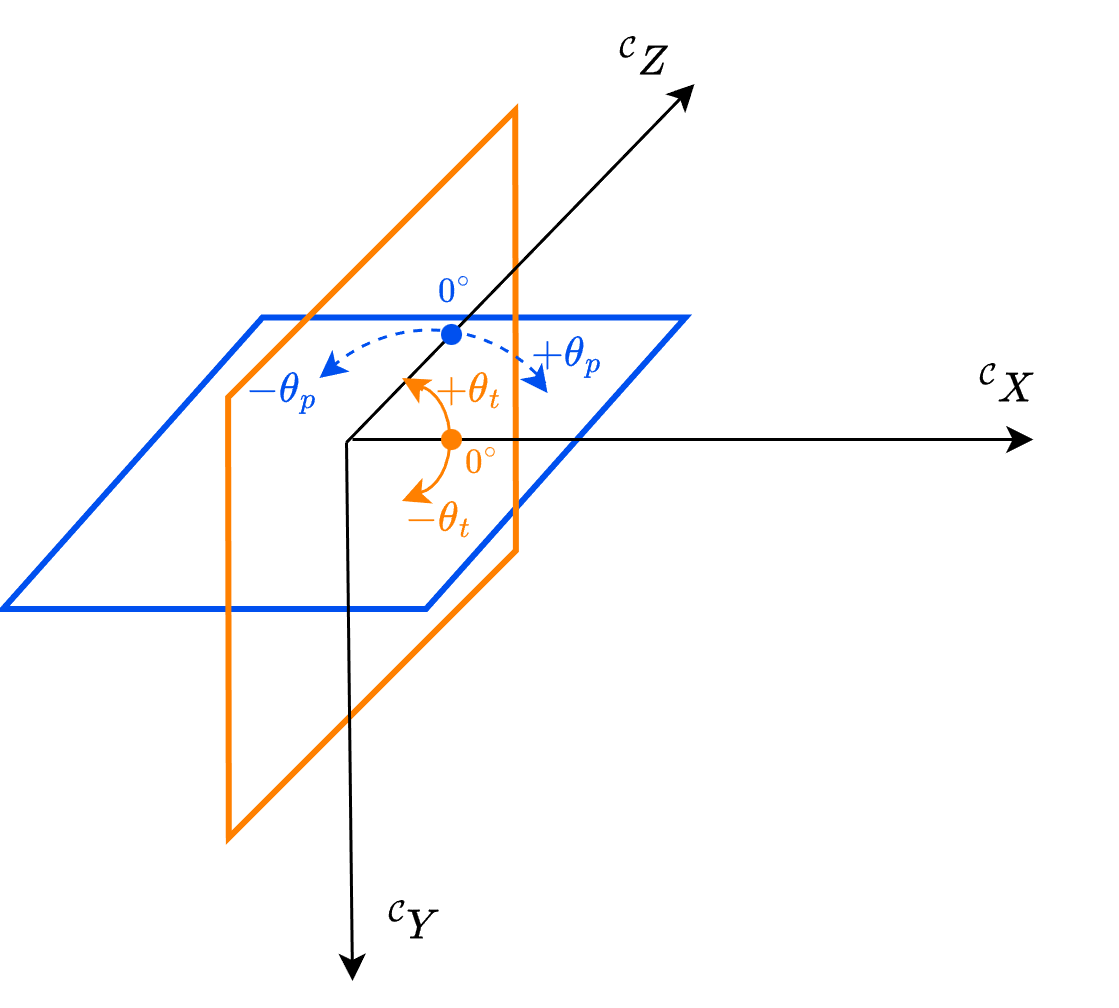}
    \caption{Definition of the coordinate system of the PT camera.}
    \label{fig:PTcoord}
\end{figure}

The rotational directions of the PT camera are defined in Fig. \ref{fig:PTcoord}: a positive pan angle corresponds to clockwise rotation, and a positive tilt angle corresponds to upward rotation. Here, $\theta_p$ and $\theta_t$ denote the pan and tilt angles, respectively.

\subsection{UKF\cite{Junming2020}}
A UKF is employed to estimate the target’s unknown motion. The UKF utilizes a dynamic model and measurements from the PT camera to determine the target's state. The nonlinear system for the UKF can be expressed as:
\begin{equation*}
    \begin{split}
        x_{k+1}&=F(x_k)+w_k\\
        z_k&=H(x_k)+v_k,
    \end{split}
\end{equation*}
where $x_k$ is the state of the target system at time $k$, $z_k$ is discrete measurement by camera at time $k$, $F(x_k)$ is a dynamics model for prediction statem $H(x_k)$ is the measurement model for correcting the predicted state, and $w_k$ and $v_k$ are process and measurement noises respectively.

The dynamics model $x\in\mathbb{R}^8$ used to predict the target motion is involved with the fixed-wing UAV, PT camera and and the tracking target, and it is defined as:
\begin{equation}
    x=\begin{bmatrix}x_1&x_2&^\mathcal{G}r^T_q&^\mathcal{G}V^T_q\end{bmatrix}^T.
    \label{eq:UKFstate}
\end{equation}
where $x_1$ and $x_2$ are defined as follow:
\begin{equation*}
\begin{bmatrix}x_1&x_2\end{bmatrix}=\begin{bmatrix}\frac{X}{Z}&\frac{Y}{Z}\end{bmatrix},
\end{equation*}
where
\begin{equation*}
    \begin{matrix}
    {^\mathcal{C}r}_{q/c}=[{^\mathcal{C}x_{q/c}}\ ^\mathcal{C}y_{q/c}\ ^\mathcal{C}z_{q/c}]^T=[X\ Y\ Z]^T.\\
    \end{matrix}
\end{equation*}
Assuming that the target follows a constant-velocity motion, the time derivative of the system state in \eqref{eq:UKFstate} is given by
\begin{equation}
    \dot{x} =\begin{bmatrix}
        \dot x_1\\\dot x_2\\ \dot{^\mathcal{G}r^T_q}\\ \dot{^\mathcal{G}V^T_q}
    \end{bmatrix}=
\begin{bmatrix}
v_{qx} x_3 - v_{qz} x_1 x_3 + \zeta_1 + \Omega_1 + \eta_1 \\
v_{qy} x_3 - v_{qz} x_2 x_3 + \zeta_2 + \Omega_2 + \eta_2 \\
\dot{r}_{q/c} + {V}_c + \boldsymbol{\omega}_c \times {r}_{q/c} \\
{0}_{3 \times 1}
\end{bmatrix}
\label{eq:6},
\end{equation}
where
\begin{equation}
    \begin{split}
        x_3 &= \frac{1}{Z}\\
\zeta_1 &= \omega_{cx} x_1 x_2 - \omega_{cy} - \omega_{cy} x_1^2 + \omega_{cx} x_2\\
\zeta_2 &= \omega_{cx} + \omega_{cx} x_2^2 - \omega_{cy} x_1 x_2 - \omega_{cx} x_1\\
\Omega_1 &= \omega_{px} x_1 (x_2 + m_y x_3) - \omega_{py} - \omega_{py} x_1^2 \\&- \omega_{py} x_3 (m_z + m_x x_1) + \omega_{pz} (x_2 + m_y x_3)\\
\Omega_2 &= \omega_{pz} + \omega_{px} x_3 (m_z + m_y x_2) + \omega_{px} x_2^2 \\&- \omega_{py} x_2 (x_1 + m_x x_3) + \omega_{pz} (-x_1 - m_x x_3)\\
\eta_1 &= (v_{px} x_1 - v_{px}) x_3\\
\eta_2 &= (v_{px} x_2 - v_{py}) x_3,
    \end{split}
    \label{eq:8}
\end{equation}
and $\boldsymbol{\omega}_c = [\omega_{cx}, \omega_{cy}, \omega_{cz}]^T$ denotes the angular velocity vector associated with the camera’s motion, all the coordinates in (\ref{eq:8}) are expressed in the camera frame $\mathcal{C}$ and $r_{c/p}=\begin{bmatrix}
    m_x&m_y&m_z
\end{bmatrix}^T$.

The measurements used by the UKF consist of the target’s 2D image coordinates and the camera position, expressed as
\begin{equation}
\mathbf{z} =
\begin{bmatrix}
u & v & \mathbf{r}_c^{T}
\end{bmatrix}^{T}.
\label{eq:9}
\end{equation}

Target detection in the image is performed using the You Only Look Once (YOLO) algorithm. After training, YOLO outputs a bounding box around the detected target, and the center of this box is selected as the image feature vector $Q_t(u,v)$, as illustrated in Fig.~\ref{fig:framework}.

Based on the pinhole camera model, the normalized image coordinates $x_1$ and $x_2$ are given by
\begin{equation}
\begin{split}
x_1 &= \frac{u - c_u}{f_x}, \\
x_2 &= \frac{v - c_v}{f_y},
\end{split}
\label{eq:10}
\end{equation}
where $c_u$ and $c_v$ denote the principal point of the image, and $f_x$ and $f_y$ are the focal lengths along the horizontal and vertical axes, respectively.
Based on \eqref{eq:9} and \eqref{eq:10}, the measurement model used for UKF is given by
\begin{equation*}
{z} =
\begin{bmatrix}
u \\ v \\ {r}_c
\end{bmatrix}
=
\begin{bmatrix}
f_x x_1 + c_u \\
f_y x_2 + c_v \\
{r}_q - {r}_{q/c}
\end{bmatrix},
\end{equation*}
which relates the measurement vector to the UKF state estimate.

\begin{figure}[h]
    \centering
    \includegraphics[width=\columnwidth]{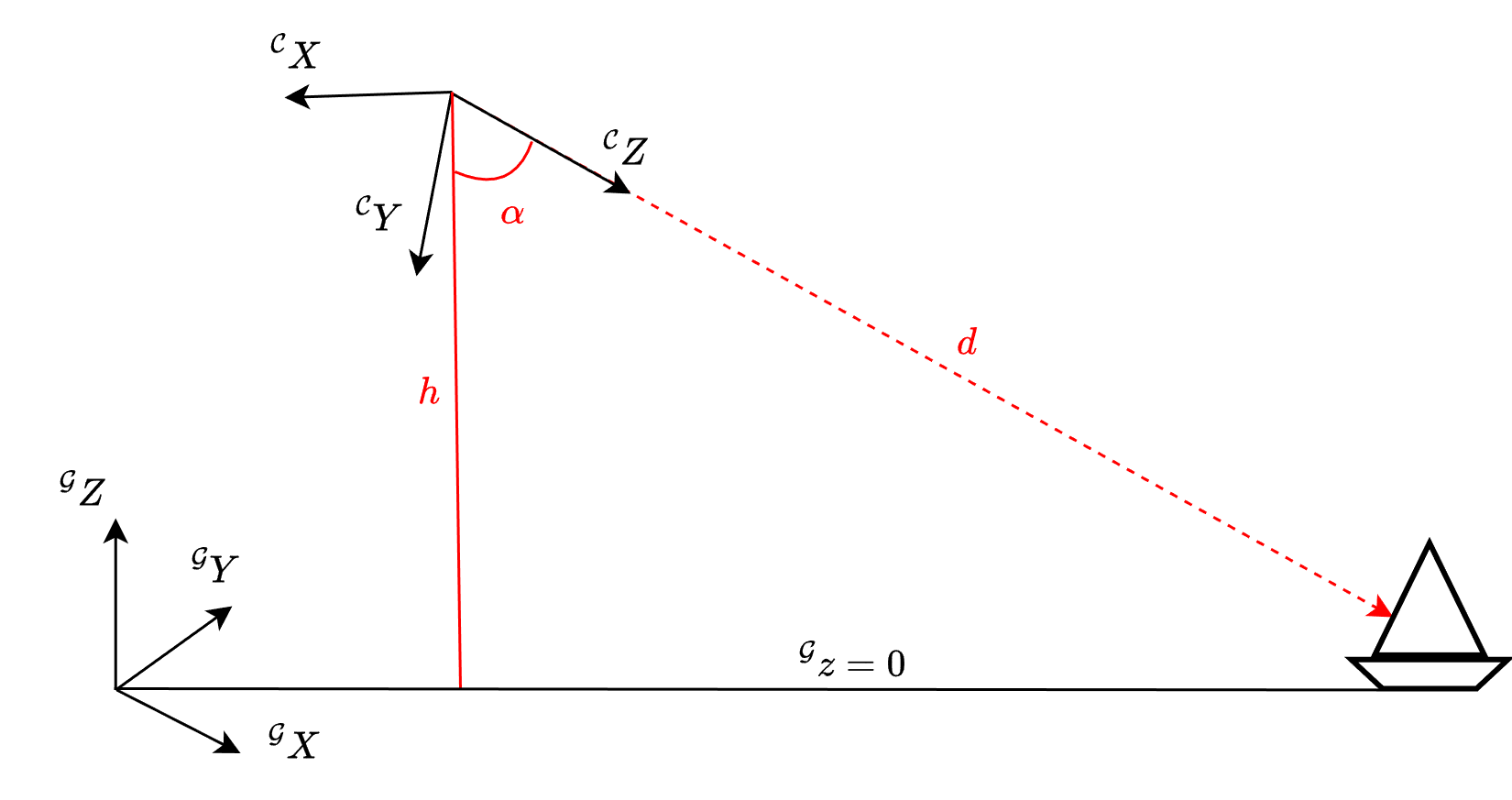}
    \caption{The relation of camera depth under the assumptions.}
    \label{fig:depth}
\end{figure}

In long-range fixed-wing UAV missions, directly determining the reciprocal of $x_3$—that is, the image depth $Z$ in \eqref{eq:8}—is challenging. Therefore, depth must be estimated using additional measurements. 

One of the depth estimation approaches relies on geometric relationships. Specifically, the geometry used for depth computation is shown in Fig.~\ref{fig:depth}. The depth can be derived from trigonometric relations as
\begin{equation*}
\begin{split}
\alpha &= \cos^{-1}(- {e}_3^{T} {c}_3 ), \\
d &= Z = \frac{1}{x_3} = \frac{h}{\cos \alpha},
\end{split}
\end{equation*}
where ${e}_3 \in \mathbb{R}^3$ denotes the unit vector aligned with the inertial $^{\mathcal{G}}Z$ axis,  ${c}_3 \in \mathbb{R}^3$ represents the unit vector aligned with the camera $^{\mathcal{C}}Z$ axis, and the camera height $h$ is determined from the UAV’s altitude.
\section{Control System Design}
In this section, several controllers are designed to accomplish the full tracking-to-terminal mission. Preserving visual contact with the target is essential in all mission phases. In Phase 1, the IBVS controller locks onto the target image to generate an initial estimate while the UAV follows a simple flight trajectory. In Phases 2 and 3, it continuously keeps the target centered in the camera FOV, thereby enabling robust tracking, complying with UAV dynamics and PT motion constraints, and facilitating accurate terminal guidance. This section details the design of each controller and explains its function within the three consecutive mission phases illustrated in Fig.~\ref{fig:phases}.

\subsection{Phase 1: Controller design of the PT camera System}
This section presents the design of the IBVS controller, which operates across all flight phases. The IBVS controller generates the required camera rotation commands to maintain the target at the center of the image, and these commands are then converted by the PT controller so that the PT rotations are controlled in accordance with the IBVS controller output for the camera attitude.
\subsubsection{IBVS controller}
The feature vector $s(t)=\begin{bmatrix}
    x_1, &x_2
\end{bmatrix}^T$ is defined as the control state, where $x_1$ and $x_2$ is same defined as \eqref{eq:UKFstate}. The control objective for IBVS is to maintain the target in the center of the camera image, the desired state can be expressed:
\begin{equation*}
    s^*=\begin{bmatrix}
        0&0
    \end{bmatrix}^T.
\end{equation*}
The error between the current and desired state can be expressed:
\begin{equation}
    e(t)=s(t)-s^*
    \label{eq:11},
\end{equation}
where $e(t): [0, \infty) \to \mathbb{R}^2$ and $s^*$ denotes a given constant element, differentiating (\ref{eq:11}) yields:
$$\dot{e}(t)=\dot{s}(t).$$
By incorporating the system's dynamic model defined in (\ref{eq:6}), the system equation can be expressed as:
\begin{equation}
    \dot{e}=\begin{bmatrix}\dot{x}_1\\\dot{x}_2\end{bmatrix}=L_1\boldsymbol{\omega}_c+L_2\begin{bmatrix}V_p-V_q\\\omega_p\end{bmatrix}
    \label{eq:IBVSerror},
\end{equation}
where $\boldsymbol{\omega}_c = [\omega_{cx}, \omega_{cy}, \omega_{cz}]^T$ denotes the angular velocity vector associated with the camera’s motion, and $L_1$ denotes the Jacobian matrix that relates the  angular velocity of the camera frame to the feature vector, defined as
$$L_1=\begin{bmatrix}x_1x_2&-(1+x_1^2)&x_2\\1+x_2^2&-x_1x_2&-x_1\end{bmatrix},$$
and 
$L_2$ is an interaction matrix defined in (\ref{eq:L2}).

\begin{figure*}[t] 
\normalsize 
\begin{equation}
L_2 = 
\begin{bmatrix}
-x_3 & 0 & x_1 x_3 & x_1(x_2+m_y x_3) & -(1+x_1^2 + x_3(m_z+m_x x_1)) & x_2 + m_y x_3 \\
0 & -x_3 & x_2 x_3 & 1 + x_2^2 + x_3(m_z + m_y x_2) & -x_2(x_1+m_x x_3) & -x_1 - m_x x_3
\end{bmatrix}
\label{eq:L2},
\end{equation}
\hrulefill 
\end{figure*}

The resulting closed-loop error dynamics can be expressed as:
\begin{equation*}
    \dot{e}=-\lambda e,
\end{equation*}
where $\lambda \in \mathbb{R}^+$ is the control gain. By reducing the camera angular velocity $\boldsymbol{\omega}_c$ to the two PT degrees of freedom, the IBVS controller can be designed from the error dynamics in \eqref{eq:IBVSerror} as
\begin{equation}
    \begin{bmatrix}\omega_{cx}\\\omega_{cy}\end{bmatrix}=-\hat{L}_1^{-1}L_2\begin{bmatrix}V_p-V_q\\\omega_p\end{bmatrix}-\lambda\hat{L}_1^{-1}e,
    \label{eq:IBVSsss}
\end{equation}
where $\hat{L}_1$ is a submatrix of $L_1$ defined as:
$$\hat{L}_1=\begin{bmatrix}x_1x_2&-(1+x_1^2)\\1+x_2^2&-x_1x_2\end{bmatrix}.$$
\subsubsection{PT camera controller}
Following the pan-then-tilt rotation order as shown in Fig. \ref{fig:Gimbalcoord}, the required PT angular velocity \eqref{eq:IBVSsss} of the PT camera can be derived from the camera angular velocity defined in (\ref{eq:IBVSsss}) as
\begin{equation}
\resizebox{1\columnwidth}{!}{$
\boldsymbol{\omega}_c=\begin{bmatrix}0&1&0\\0&0&1\\1&0&0\end{bmatrix}\begin{bmatrix}\cos\theta_t&0&-\sin\theta_t\\0&1&0\\\sin\theta_t&0&\cos\theta_t\end{bmatrix}\begin{bmatrix}0\\0\\\dot{\theta}_p\end{bmatrix}+\begin{bmatrix}0&1&0\\0&0&1\\1&0&0\end{bmatrix}\begin{bmatrix}0\\\dot{\theta}_t\\0\end{bmatrix}.
    \label{eq:omegac}$}
\end{equation}
The resulting pan and tilt rates can then be computed from \eqref{eq:omegac} as
\begin{equation}
    \begin{split}
        \dot{\theta_t}&=\omega_{cx}\\ \dot{\theta}_p&=\frac{\omega_{cy}}{\cos\theta_t}.
    \end{split}
    \label{eq:sigular}
\end{equation}
It is worth noting that a singularity occurs when $\theta_t = \pm 90^\circ$. This issue will be addressed in coordination with other controllers.

\subsection{Phase 2: NMPC for the tracking phase}

\begin{figure}[h]
    \centering
    \includegraphics[width=\columnwidth]{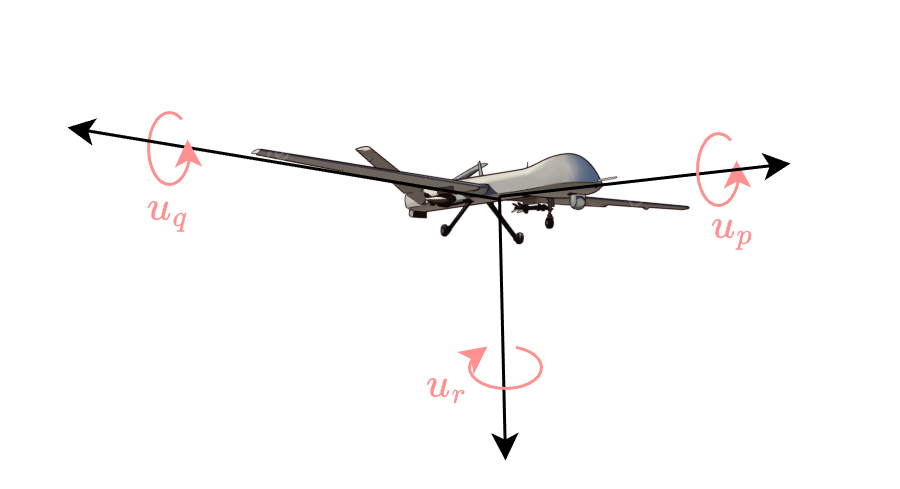}
    \caption{Definition of the body-frame angular velocity control inputs.}
    \label{fig:upqr}
\end{figure}

During Phase 2, the NMPC is employed to control the UAV's trajectory while respecting its dynamic constraints and the PT camera's physical limits. NMPC is a widely used optimal control strategy for nonlinear systems with constraints, making it particularly suitable for UAV tracking missions where the vehicle must follow a moving target while maintaining feasible control inputs. The NMPC system state is defined as
\begin{equation*}
    X=\begin{bmatrix}
        ^\mathcal{G}r^T_p&\phi&\theta&\psi
    \end{bmatrix}^T,
    \label{eq:NMPCstate}
\end{equation*}
where $^\mathcal{G}r^T_p$ denotes the UAV position in the inertial frame, and $\phi$, $\theta$, and $\psi$ are same defined as \eqref{eq:1}. The control input is 
\begin{equation}
    U=\begin{bmatrix}
        V_U&u_p&u_q
    \end{bmatrix}^T,
    \label{eq:NMPCinput}
\end{equation}
where $V_U$ is the forward velocity commanded along the body-frame $^\mathcal{B}X$ axis, and $u_p$ and $u_q$ are the angular velocities along $^\mathcal{B}X$ and $^\mathcal{B}Y$, respectively, as shown in Fig.~\ref{fig:upqr}. 
By neglected the body-frame yaw rate (i.e., $u_r\approx 0$), the relationship between the body angular velocities and the Euler angle rates is given by \cite{valencia2012small}
\begin{equation}
    \begin{bmatrix}
        \dot{\phi}\\ \dot{\theta}
    \end{bmatrix}
    =
    \begin{bmatrix}
        1&\sin\phi\tan\theta\\
        0&\cos\phi
    \end{bmatrix}
    \begin{bmatrix}
        u_p\\u_q
    \end{bmatrix}.
    \label{eq:add_tran}
\end{equation}
Incorporating the translational velocity defined in (\ref{eq:2}), the angular velocity described in (\ref{eq:add_tran}) and the turning condition governed by the yaw rate as described in \cite{stevens2015aircraft}, the system dynamics are expressed as follows:
\begin{equation}
    \dot{X}=\begin{bmatrix}V_U\cos{\psi}\cos{\theta} \\
V_U\sin{\psi}\cos{\theta}\\
V_U\sin{\theta} \\
u_p+u_q\sin\phi\tan\theta\\
u_q\cos\phi\\
-\dfrac{g}{V_{U}} \tan\phi\cos\theta \end{bmatrix},
\label{eq:dynamics}
\end{equation}
where $g$ represents the gravitational constant. Given the system dynamics in \eqref{eq:dynamics}, the future state $X_{k+1}$ can be predicted from the current state $X_k$ and the control input $U_k$ as
\begin{equation*}
X_{k+1}(X_k,U_k) = X_k + \dot{X}_k T_s,
\end{equation*}
where $T_s$ is the sampling time. Accordingly, for a given current state $X_k$ and a sequence of control inputs $\mathcal{U} = \begin{bmatrix} U_k, U_{k+1}, \ldots, U_{k+N_c} \end{bmatrix}$, a corresponding sequence of predicted future states $\mathcal{X} = \begin{bmatrix} X_{k+1}, X_{k+2}, \ldots, X_{k+N_p} \end{bmatrix}$ can be obtained. Here, $N_p$ and $N_c$ denote the prediction and control horizons, respectively. In this study, the prediction and control horizons are set as the same, $N_p=N_c=N$.
\subsubsection{NMPC}
An optimal cost function $J(\mathcal{X}, \mathcal{U})$ is defined over the predicted state sequence $\mathcal{X}$. To accomplish the tracking objective in Phase~2, the cost function consists of four components,
\begin{equation*}
J(\mathcal{X}, \mathcal{U}) =
\sum_{k=0}^{N-1}
\big( f_1(X_k) + f_2(X_k) + f_3(X_k) + f_4(U_k) \big),
\end{equation*}
where each term addresses a specific tracking requirement.
The term $f_1(X_k)$ regulates the planar distance between the UAV and the target on the inertial $XY$-plane, $$f_1(X_k)=\begin{Vmatrix}\sqrt{(^\mathcal{G}x_{p,k}-{^\mathcal{G}x_{q,k}})^2+(^\mathcal{G}y_{p,k}-{^\mathcal{G}y_{q,k}})^2}-R \end{Vmatrix}^2_{w_1},$$
    where ${}^{\mathcal G} r_{p,k} = [{}^{\mathcal G}x_{p,k}\; {}^{\mathcal G}y_{p,k}\; {}^{\mathcal G}z_{p,k}]^T$
and ${}^{\mathcal G} r_{q,k} = [{}^{\mathcal G}x_{q,k}\; {}^{\mathcal G}y_{q,k}\; {}^{\mathcal G}z_{q,k}]^T$
denote the positions of the UAV and the target expressed in the inertial frame $\mathcal G$, respectively.
$f_1$ drives the UAV to keep a target separation of $R$, ensuring both dependable visual tracking and advantageous terminal geometry. The altitude control component is given by
$$f_2(X_k)=\begin{Vmatrix}^\mathcal{G}z_{p,k}-H \end{Vmatrix}^2_{w_2},$$
where $H$ is the target flight altitude and $w_2$ is the weight associated with this objective. The term $f_2$ imposes a penalty on any deviation from the specified flight altitude. To limit aggressive maneuvers and decrease control effort, an attitude smoothness term is introduced as
$$f_3(X_k)=\begin{Vmatrix}Q_{k+1}-Q_k \end{Vmatrix}^2_{w_3},$$
where $Q_k=\begin{bmatrix}\phi_{k}& \theta_{k} & \psi_{k} \end{bmatrix}^T$ and $w_3\in\mathbb{R}^{3\times3}$ is a diagonal weighting matrix. The function $f_3$ penalizes changes in the Euler angles between successive sampling instants. Finally, continuity of the input across successive NMPC horizons is encouraged through
$$f_4=\begin{Vmatrix}
    \begin{bmatrix}
        u^*_{p,-1}\\u^*_{q,-1}
    \end{bmatrix}-
    \begin{bmatrix}
        u_{p,0}\\u_{q,0}
    \end{bmatrix}
\end{Vmatrix}^2_{w_4},$$
where $u^*_{p,-1}$ and $u^*_{q,-1}$ are the optimal body rates applied to the UAV in the previous NMPC solution, and $u_{p,0}$ and $u_{q,0}$ denote the first input elements of the current control horizon.
To compute the optimal control sequence $\mathcal{U}^*,$ the NMPC problem is formulated by minimizing the cost function $J(\mathcal{X},\mathcal{U})$, subject to state and input constraints,
\begin{equation}
\begin{aligned}
\mathcal{U}^*
&= \arg\min_{\mathcal{U}} \; J(\mathcal{X},\mathcal{U}) \\
\text{s.t.}\quad
& \mathcal{X}_{lb} \le \mathcal{X} \le \mathcal{X}_{ub}, \\
& \mathcal{U}_{lb} \le \mathcal{U} \le \mathcal{U}_{ub},
\end{aligned}
\label{eq:NMPC}
\end{equation}
where $\mathcal{X}_{lb}$ and $\mathcal{X}_{ub}$ represent the lower and upper bounds of the predicted state set $\mathcal{X}$, while $\mathcal{U}_{lb}$ and $\mathcal{U}_{ub}$ denote the lower and upper bounds of the control input set $\mathcal{U}$.  In this case, they are set to match the physical limits of UAV motion. Thus, the possible UAV attitude, input velocity, and angular rate are defined as follows:
\begin{equation}
    \begin{split}
        \phi_{min}\leq&\phi_k\leq\phi_{max}\\
        \theta_{min}\leq&\theta_k\leq\theta_{max}\\
        V_{U,min}\leq &V_{U,k}\leq V_{U,max}\\
        u_{p,min}\leq&u_{p,k}\leq u_{p,max}\\
        u_{q,min}\leq&u_{q,k}\leq u_{q,max}.
    \end{split}
    \label{eq:limits}
\end{equation}
\begin{figure}[h]
    \centering
    \includegraphics[width=\columnwidth]{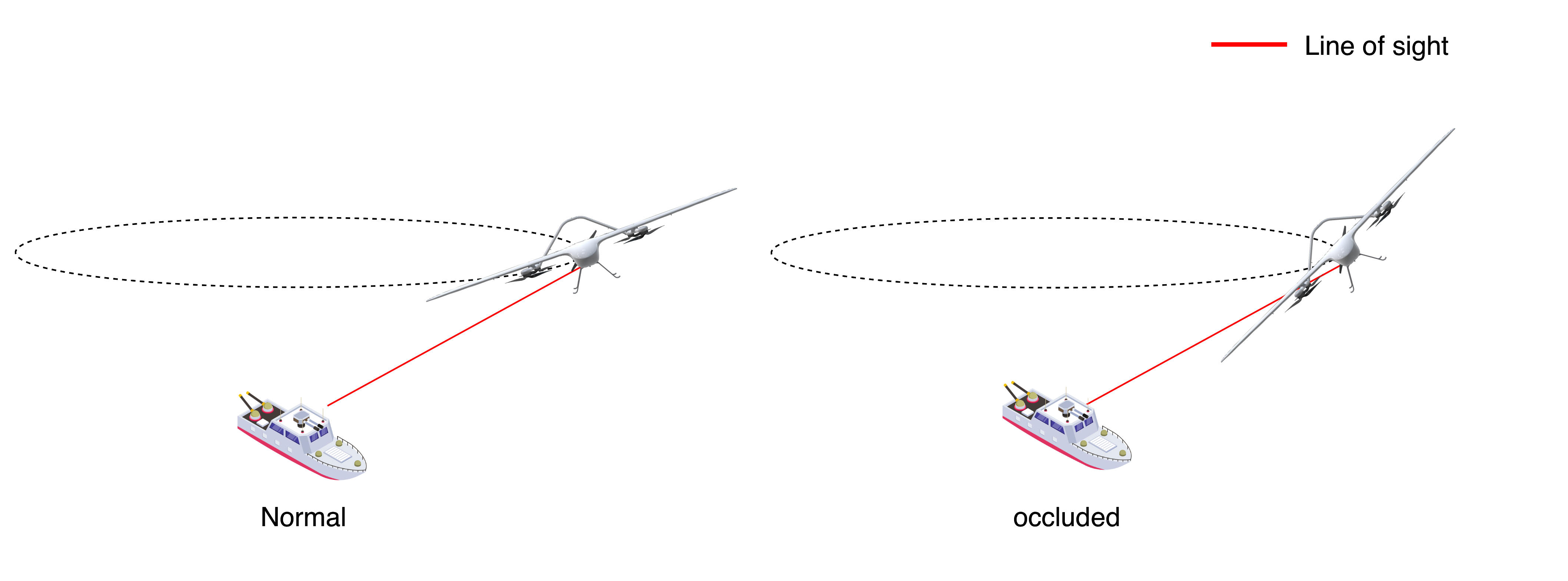}
    \caption{Illustration of UAV self-occlusion during a loitering mission: (left) normal unobstructed LOS, and (right) LOS obstruction caused by airframe-shadowing during a banking turn.}
    \label{fig:sight_block}
\end{figure}
\subsubsection{To prevent self-occlusion} \label{sec:self-occlusion}
\begin{figure}[h]
    \centering
    \includegraphics[width=\columnwidth]{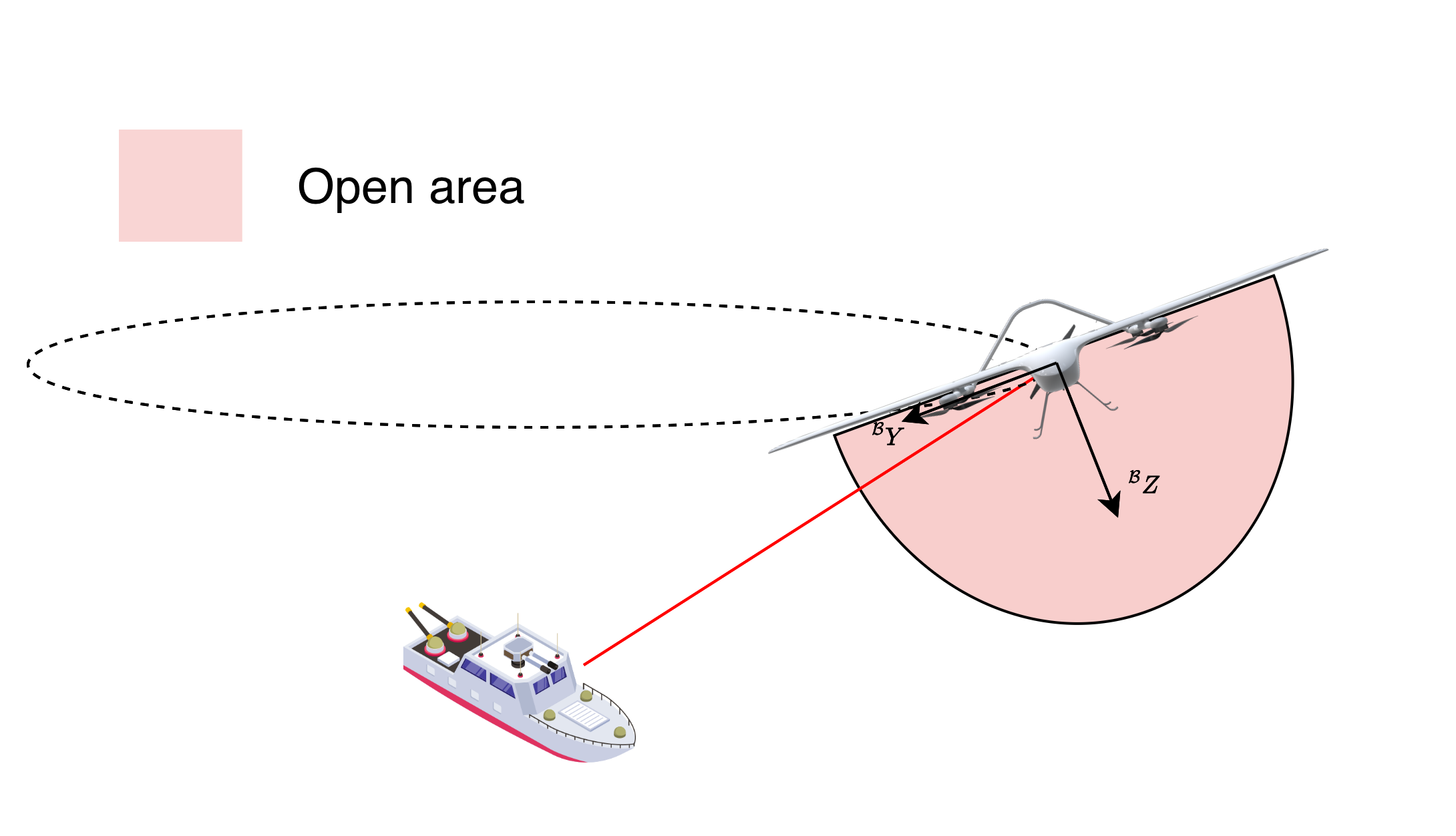}
    \caption{}
    \label{fig:reachable}
\end{figure}

In addition to (\ref{eq:limits}), an extra motion constraint on the UAV bank angle is imposed to avoid visual self-occlusion, as depicted in Fig.~\ref{fig:sight_block}, assuming that the tilt angle is constrained to nonpositive values in accordance with the mechanical limits shown in Fig.~\ref{fig:PTcoord}. In the normal case (left side of Fig.~\ref{fig:sight_block}), the line of sight (LOS) between the onboard sensor and the maritime target remains unobstructed as long as the PT camera satisfy the joint limits (i.e., $-\pi\leq\theta_p\leq\pi$, $-\frac{\pi}{2}\leq\theta_t\leq0$). In contrast, as the UAV increases its bank angle to sustain circular flight, it enters the occluded scenario (right side), in which the tilted wing crosses and obstructs the LOS vector, despite the fact that the same PT camera motion constraints are still satisfied. Thus, the onset of occlusion is determined by the roll angle $\phi$, and the target remains visible and trackable as long as it lies along the positive direction of the UAV body-frame $Z$-axis, i.e., $^\mathcal{B}z\geq0$, as shown in Fig.~\ref{fig:reachable}.
The relative position between the UAV and the target in the inertial frame $\mathcal{G}$ is given by
\begin{equation*}
    ^\mathcal{G}r_{q/p}=^\mathcal{G}r_q-{^\mathcal{G}r}_p=
\begin{bmatrix}
    ^\mathcal{G}x_{q/p}&^\mathcal{G}y_{q/p}&^\mathcal{G}z_{q/p}
    \label{eq:21}
\end{bmatrix}^T.
\end{equation*}
By neglecting the UAV pitch angle (a valid assumption in the Phase 2 stage) and using the coordinate transformation given in \eqref{eq:1}, the relative position in the UAV body frame $\mathcal{B}$ can be written as
\begin{equation*}
 ^\mathcal{B}r_{q/p}=
        \begin{bmatrix}
        ^\mathcal{G}x_{q/p}\sin\psi+ {^\mathcal{G}y}_{q/p}\cos\psi\\
         ^\mathcal{G}x_{q/p}\cos\psi\cos\phi- {^\mathcal{G}y}_{q/p}\sin\psi\cos\phi- {^\mathcal{G}z}_{q/p}\sin\phi\\
         -{^\mathcal{G}x}_{q/p}\cos\psi\sin\phi+ {^\mathcal{G}y}_{q/p}\sin\psi\sin\phi- {^\mathcal{G}z}_{q/p}\cos\phi
    \end{bmatrix}.
\end{equation*}
Given the preference $^\mathcal{B}z\geq0$ illustrated in Fig.~\ref{fig:sight_block}, it is desirable for the target to lie above the UAV’s body-frame XY-plane (${}^\mathcal{B}z_{q/p}\geq0$). Accordingly, the constraint is formulated as
\begin{equation*}
    ^\mathcal{B}z_{q/p}=-^\mathcal{G}x_{q/p} \cos\psi \sin\phi + {^\mathcal{G}y}_{q/p} \sin\psi \sin\phi - {^\mathcal{G}z}_{q/p} \cos\phi \geq 0,
\end{equation*}
which can be expressed as
\begin{equation}
\tan\phi \;
\begin{cases}
\le -\,{}^\mathcal{G}z_{q/p}/R_p, & R_p > 0,\\
\ge -\,{}^\mathcal{G}z_{q/p}/R_p, & R_p < 0,
\end{cases}
\label{eq:constraint}
\end{equation}
where $R_p = {}^\mathcal{G}x_{q/p}\cos\psi - {}^\mathcal{G}y_{q/p}\sin\psi.$

This constraint guarantees that the target remains within the camera’s field of view while also enforcing that every mechanically reachable target lies above the UAV’s body-frame XY-plane. 
To additionally guarantee that the constraints in (\ref{eq:constraint}) are fulfilled via the rolling $u_p$ in the NMPC as defined in (\ref{eq:NMPC}), control barrier functions (CBFs) are employed to recast these constraints into inequality form, which will be merged into the constraints in (\ref{eq:limits}). We first examine the situation where, according to \eqref{eq:constraint}, the target is positioned on the right-hand side of the UAV, which is described by
\begin{equation*}
    R_p \geq 0.
\end{equation*}
To satisfy the condition stated above, we define a scalar function $h(\phi)$, and we aim to impose the following inequality constraint:
\begin{equation}
    h(\phi)
    =
    \tan^{-1}\!\left(
    \frac{-{^\mathcal{G}z}_{q/p}}
    {R_p}
    \right)
    - \phi
    \geq 0,\quad R_p \geq 0.
    \label{eq:hCBF}
\end{equation}
Accordingly, the CBF condition is defined as
\begin{equation}
    \dot h \geq -\beta h,\quad R_p \geq 0,
    \label{eq:hlarge}
\end{equation}
where the dynamics $\dot \phi$ defined in (\ref{eq:dynamics}) is used, and $\beta > 0$ is a design parameter that determines the strictness of the constraint enforcement.
By taking the time derivative of the scalar function $h(\phi)$ defined in (\ref{eq:hCBF}) and considering the CBF condition given in (\ref{eq:hlarge}), the following nonlinear inequality constraint on the control input is obtained:
\begin{equation}
\resizebox{1\columnwidth}{!}{$
    u_p
    \leq
    \beta \left(
    \tan^{-1}\!\left(
    \frac{-{^\mathcal{G}z}_{q/p}}
    {R_p}
    \right)
    - \phi
    \right)+\alpha- u_q \sin\phi \tan\theta,
    \label{eq:cbfright}
    $}
\end{equation}
for $R_p \geq 0,$ where $\alpha$ is the time derivative of the first term defined in (\ref{eq:hCBF}). Similarly, under the condition that the target is located on the left side of the UAV, the following inequality holds:
\begin{equation}
\resizebox{1\columnwidth}{!}{$
    u_p
    \geq
    \beta \left(
    \tan^{-1}\!\left(
    \frac{-{^\mathcal{G}z}_{q/p}}
    {R_p}
    \right)
    - \phi
    \right)+\alpha- u_q \sin\phi \tan\theta,
    \label{eq:cbfleft}
    $}
\end{equation}
for $R_p \leq 0.$ Based on the target's relative position, the nonlinear input constraints (\ref{eq:cbfright}) and (\ref{eq:cbfleft}) are incorporated into the NMPC formulation (\ref{eq:limits}) to guarantee that the UAV remains within the camera's tilt-angle limits.

\subsection{Phase 3: Biased proportional navigation for terminal guidance}
A BPNG scheme is utilized to impose the impact angle constraint (IAC), guaranteeing that the UAV reaches the target with a specified terminal angle. When Phase~3 begins, control authority is handed over from the NMPC to the BPNG controller, which directs the UAV toward the moving target while shaping the trajectory to fulfill the IAC and preserving visual tracking through the camera system. In this phase, the control input changes from (\ref{eq:NMPCinput}) to the UAV’s angular velocity
\begin{equation}
\omega_{{cmd}}
=\begin{bmatrix} u_p & u_q & u_r, \end{bmatrix}^T
\label{eq:steering}
\end{equation}
as depicted in Fig. \ref{fig:upqr}, to steer the UAV so that it is aimed at the target. Therefore, a quaternion-based BPNG controller, developed based on the approach in \cite{kim2021quaternion}, is adopted for terminal guidance of the controlled UAV.
The optimal three-dimensional IAC law derived from the quaternion-based BPNG framework in \cite{kim2021quaternion} is given by
\begin{equation}
    u^*_{IAC}=N_\sigma\Omega\times V_p+\frac{N_fV_p}{t_{go}}\times(\frac{V_p\times V_f}{|V_p\times V_f|})\cos^{-1}(\frac{V_p\cdot V_f}{|V_p||V_f|}),
    \label{eq:final_IAC}
\end{equation}
where $V_p$ denotes the UAV velocity direction vector, which is assumed to be aligned with the body $x$-axis under coordinated flight with negligible sideslip, $V_f$ is the desired terminal velocity direction corresponding to the prescribed impact angle, and $t_{go}$ is the estimation time-to-go until impact, defined as
\begin{equation*}
    t_{go}=\frac{|r_{q/p}|}{|V_p|},
\end{equation*}
with $r_{q/p}$ representing the relative position vector between the UAV and the target. The guidance gains $N_\sigma$ and $N_f$ are selected based on the optimal values reported in \cite{kim2021quaternion}, and are set to $N_\sigma=6$ and $N_f=2$ in this study.
\begin{figure}
    \centering
    \includegraphics[width=0.75\linewidth]{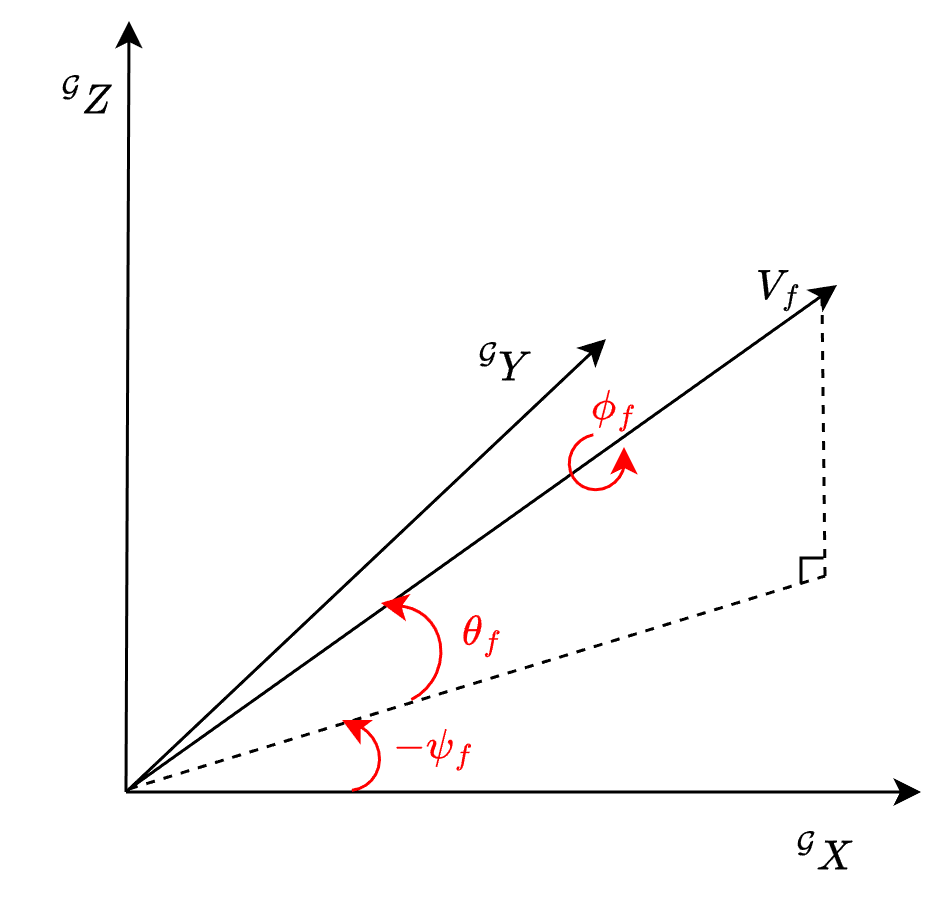}
    \caption{The relation between terminal angles and terminal velocity vector.}
    \label{fig:impact_ang}
\end{figure}
The relationship between the desired terminal velocity direction $V_f$ and the prescribed impact pitch and yaw angles $\theta_f$ and $\psi_f$, as illustrated in Fig.~\ref{fig:impact_ang}, is given by
\begin{equation*}
    V_f=\begin{bmatrix}
        \cos\theta_f\cos\psi_f,&\cos\theta_f\sin\psi_f,&-\sin\theta_f
        \end{bmatrix}^T,
\end{equation*}
where the impact roll angle does not influence the terminal velocity direction and is therefore neglected. Under the coordinated-flight assumption, the optimal control input obtained from \eqref{eq:final_IAC} is always perpendicular to the UAV velocity vector. Consequently, when expressed in the UAV body frame, the IAC input can be decomposed solely along the $^\mathcal{B}Y$ and $^\mathcal{B}Z$ axes as
\begin{equation}
\begin{split}
    ^\mathcal{B}u&=R_x(\phi)R_y(\theta)R_z(\psi)u^*_{IAC}\\
    &=\begin{bmatrix}
        0&u_y&u_z
    \end{bmatrix}^T.
\end{split}\label{eq:IAC}
\end{equation}
The commanded body-rate vector specified in (\ref{eq:steering}) can be determined from the centripetal acceleration experienced during navigation
\begin{equation}
    ^\mathcal{B}u=\omega_{cmd}\times {^\mathcal{B}V_p}=\begin{bmatrix}
        0\\ u_rV_U\\ -u_qV_U
    \end{bmatrix},
    \label{eq:bpngdot}
\end{equation}
where $^{\mathcal{B}}V_{p} = \begin{bmatrix} V_{U} & 0 & 0 \end{bmatrix}$
under the assumption of coordinated flight with negligible sideslip.
By setting the resulting acceleration in (\ref{eq:IAC}) equal to the commanded acceleration in \eqref{eq:bpngdot}, we obtain:
\begin{equation}
    \omega_{cmd}  =\begin{bmatrix}
        0\\ -\frac{u_z}{V_U}\\ \frac{u_y}{V_U}
    \end{bmatrix},
    \label{eq:bpng_cmd}
\end{equation}
These are then sent to the flight controller as the ultimate control commands, while $u_p$ is left uncontrolled because it has no influence on aiming.

\subsection{Control Architecture}
\begin{figure}
    \centering
    \includegraphics[width=\columnwidth]{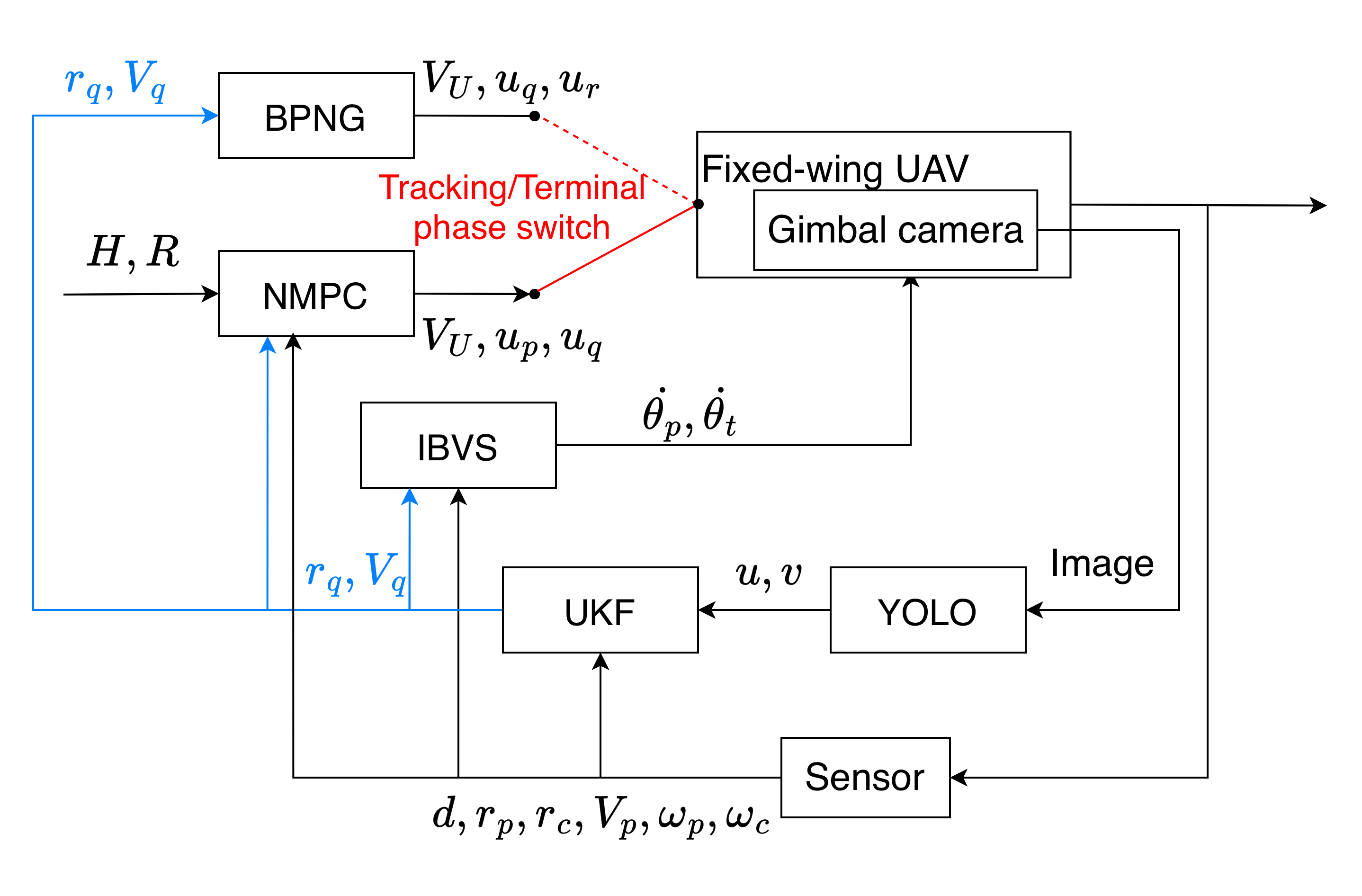}
    \caption{Block diagram of the Control architecture.}
    \label{fig:Block}
\end{figure}
The integrated control architecture for Phase 2 and Phase 3 is illustrated in the block diagram in Fig.~\ref{fig:Block}. As the foundational perception layer, the onboard sensor suite provides the UAV's states, including $d, r_p, r_c, V_p, \omega_p, \omega_c$. Concurrently, the PT camera captures the target's image, which is processed by the YOLO algorithm to detect the image-plane target location $u, v$. These visual and inertial measurements are fused via the UKF to estimate the target's position $r_q$ and velocity $V_q$. 

Based on these estimations, the framework operates in two distinct guidance phases via a Tracking/Terminal phase switch:
\begin{enumerate}
    \item \textbf{Phase 2 (Loitering Phase):} The NMPC takes $r_q$ and $V_q$, together with the  altitude $H$ and radius $R$, as inputs to compute the UAV control commands $V_U, u_p, u_q$, according to the design in (\ref{eq:NMPC}). Simultaneously, the IBVS controller regulates the PT rates $\dot{\theta}_p, \dot{\theta}_t$ using the target feedback to keep the target centered within the camera's field of view.
    \item \textbf{Phase 3 (Terminal Engagement Phase):} Once the prescribed loitering condition is satisfied, the system switches to Phase 3. The BPNG controller then becomes active, using $r_q$ and $V_q$ to generate the commands $V_U, u_q,$ and $u_r$ as specified in (\ref{eq:bpng_cmd}), thereby guiding the fixed-wing UAV toward the moving target to realize the intended impact configuration.
\end{enumerate}
\section{Simulation}
To validate the proposed methods developed in this work, real-time simulations were conducted using the Gazebo simulator \cite{koenig2004design}, which is integrated with the Robot Operating System (ROS Noetic, running on Ubuntu 20.04) and the PX4 autopilot firmware (version 1.13.2).

All simulations\footnote{https://youtu.be/m4xUBHrImvo} were performed in a three-dimensional setting to assess the performance of the proposed integrated guidance, control, and estimation framework. A fixed-wing UAV equipped with a pan–tilt camera was modeled to carry out the tracking-to-terminal mission, and a separate three-dimensional surface vehicle model served as the moving target. The specifications of both the UAV and the surface vehicle used in the simulations are provided in Table \ref{tab:model_specs}. This setup allowed for a thorough evaluation of the proposed method, confirming its effectiveness and robustness in dynamic, physically consistent environments.

\begin{table}[H]
    \centering
    \caption{Gazebo Model Specifications.}
    \renewcommand{\arraystretch}{1.2}
    \small 
    \begin{tabularx}{\columnwidth}{|l|X|r|}
        \hline
        \textbf{Model} & \textbf{Description} & \textbf{Value} \\ \hline
        \multirow{3}{*}{Fixed-wing UAV} 
            & Fuselage length (m) & 1 \\ \cline{2-3} 
            & Wingspan (m) & 2.4 \\ \cline{2-3} 
            & Wing area (m$^2$) & 0.35 \\ \hline
        \multirow{3}{*}{Water Vehicle} 
            & Width (m) & 25 \\ \cline{2-3} 
            & Length (m) & 55 \\ \cline{2-3} 
            & Height (m) & 25 \\ \hline
        \multirow{4}{*}{PT Camera}
            & Pan range (rad) & $-\pi$ to $+\pi$ \\ \cline{2-3} 
            & Tilt range (rad) & $-\pi/2$ to $+\pi/2$ \\ \cline{2-3}
            & Max. panning vel. (rad/s) & $2/3\pi$ \\ \cline{2-3}
            & Max. tilting vel. (rad/s) & $2/3\pi$ \\
        \hline
    \end{tabularx}
    \label{tab:model_specs}
\end{table}

\subsection{System Setup}
\begin{figure}[h]
    \centering
    \includegraphics[width=\columnwidth]{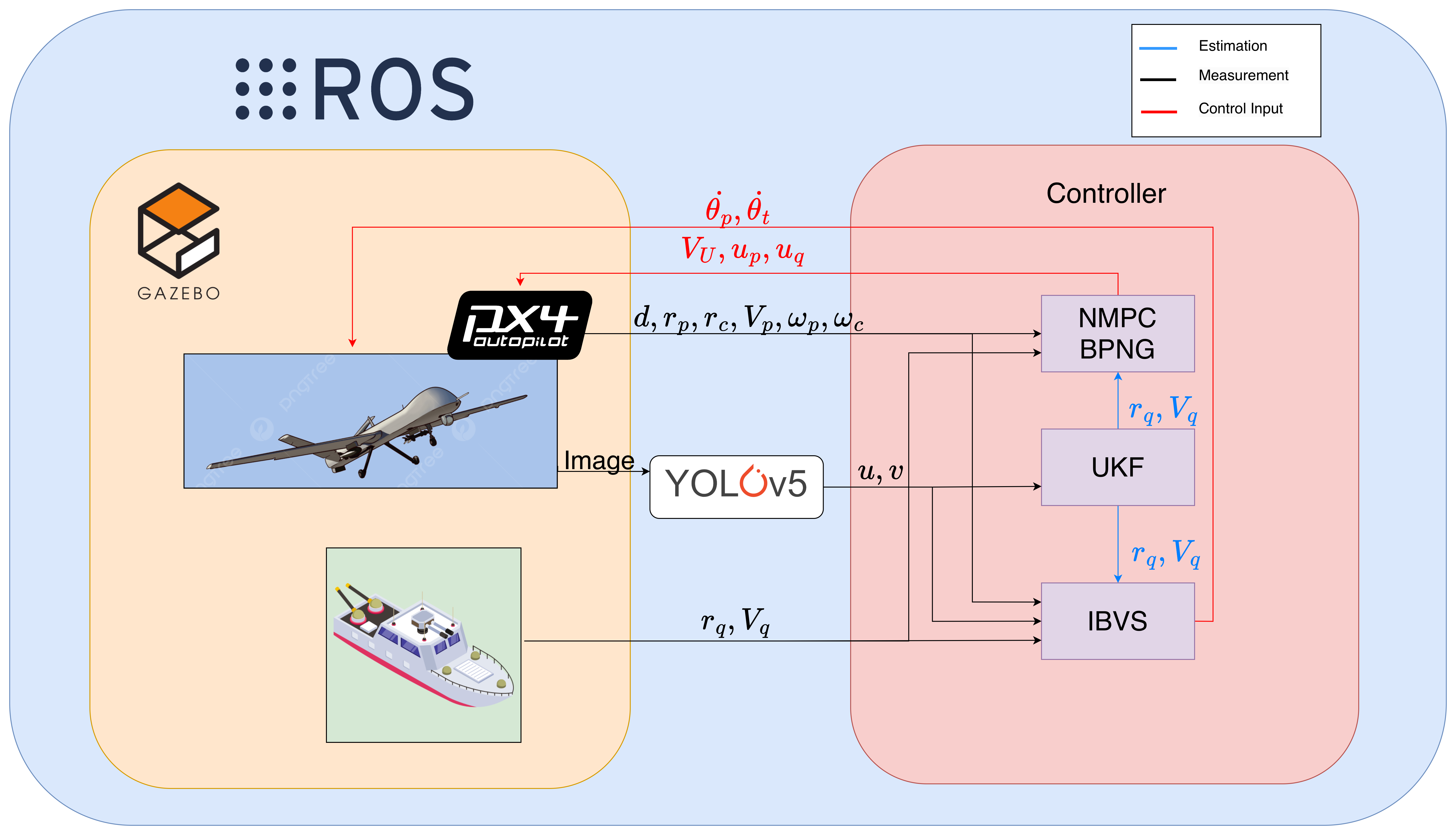}
    \caption{Simulation system architecture.}
    \label{fig:setup}
\end{figure}

The overall simulation implementation, illustrating the data flow and interaction among the UAV-target dynamics, the estimation modules, and the cooperative controllers within the ROS framework, is presented in Fig.~\ref{fig:setup}.
 The high-fidelity environment is simulated using Gazebo paired with the PX4 Autopilot, which outputs the UAV's inertial states $d, r_p, r_c, V_p, \omega_p, \omega_c$, while the simulated ship target provides its true trajectory states $r_q, V_q$ for verification. 

The mission execution follows a three-phase sequential structure:
\begin{itemize}
    \item \textbf{Phase 1 (Target Acquisition):} The UAV initially operates in Phase 1, where the PT camera searches for and firmly captures the maritime target. The raw camera images are processed by YOLOv5 to extract the image-plane target coordinates $(u, v)$, establishing a reliable visual tracking lock.
    \item \textbf{Phase 2 (NMPC-based Loitering):} Once target lock is confirmed, the framework transitions to Phase 2. The NMPC controller is configured with a prediction horizon of $N=50$ and a sampling time of $\Delta T=0.1$~s. During this phase, the UAV is commanded to loiter around the moving target with a desired planar radius of $R=500$~m while maintaining a target altitude of $500$~m, regulated via the generated commands $V_U, u_p, u_q$. Concurrently, the IBVS controller regulates the PT rates $\dot{\theta}_p, \dot{\theta}_t$ to counteract the airframe motions and mitigate potential self-occlusion risks.
    \item \textbf{Phase 3 (BPNG-based Terminal Engagement):} When both the desired loiter radius and altitude metrics are successfully satisfied, the Tracking/Terminal phase switch triggers a transition to Phase 3. The BPNG controller takes over from the NMPC to command the UAV inputs ($V_U, u_q, u_r$), steering the fixed-wing airframe toward the target to fulfill the specified terminal impact angle constraint.
\end{itemize}
In this simulation, the maritime target vehicle is modeled to move on a horizontal plane with a constant velocity vector, representing a realistic sea-surface tracking scenario. The initial position of the fixed-wing UAV is set in mid-air at $(-1000, 0, 1000)$~m, whereas the target ship departs from the coordinate origin $(0,0,0)$~m along the surface. 

Visual measurements are actively captured via the onboard PT camera, whose mechanical gimbals are subject to physical constraints: the pan angle is bounded within $[-\pi, \pi]$~rad, the tilt angle is limited within $[-\pi/2, \pi/2]$~rad. Target detection within the camera's captured frames is performed by the YOLOv5 vision module, which computes the image-plane coordinates $(u, v)$ of the target's geometric center. These noisy pixel measurements, along with the UAV's inertial states from the onboard sensors, serve as the measurement inputs to the UKF. The UKF dynamically filters the sensor noise to yield smooth, real-time target state estimations ($\hat{r}_q, \hat{V}_q$), which are subsequently distributed to the NMPC, BPNG, and IBVS control loops to close the tactical feedback.

\subsection{Trajectory and Estimation in Mission Scenario}
\begin{figure}[h]
    \centering
    \subfloat{
    \includegraphics[width=\columnwidth]{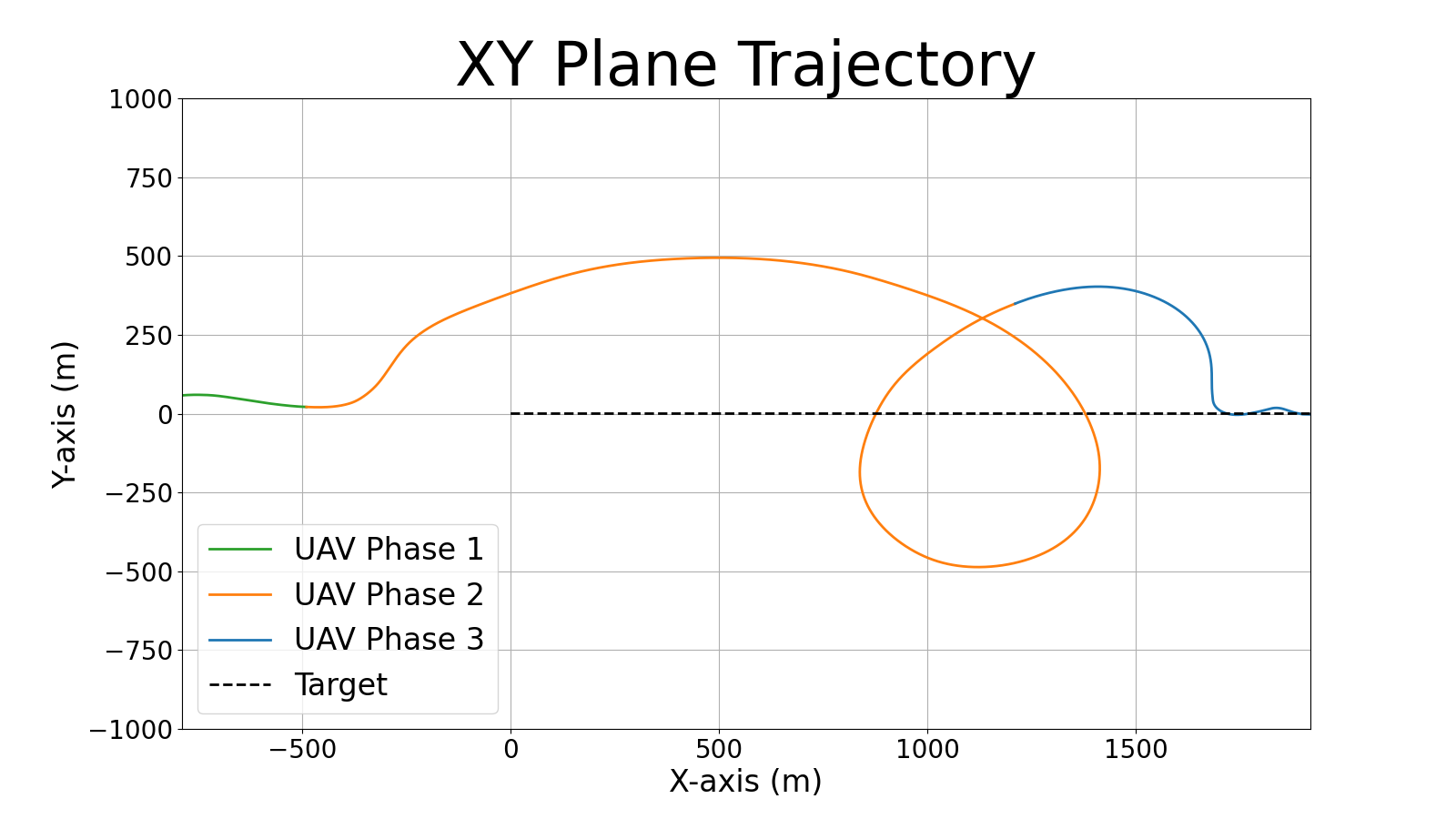}
    }\\
    \subfloat{
    \includegraphics[width=\columnwidth]{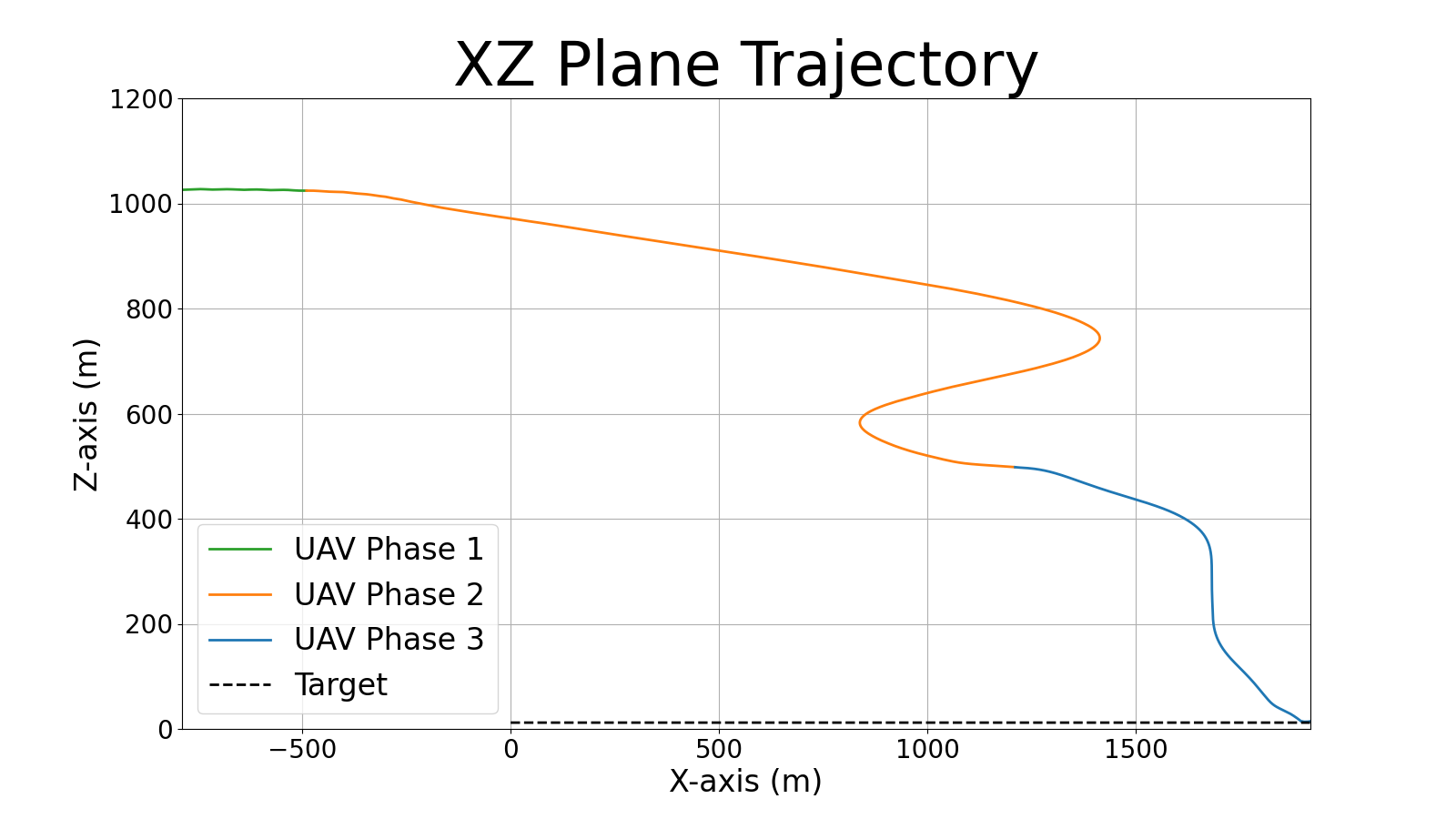}
    }
    \caption{The result tracking trajectory for fixed-wing UAV.}
    \label{fig:2Dplane}
\end{figure}

The complete three-dimensional flight trajectory of the fixed-wing UAV across all operational phases is analyzed in Fig.~\ref{fig:2Dplane}, sweeping across both the horizontal XY-plane and the vertical XZ-plane. 

\begin{itemize}
    \item \textbf{Phase 1 (Target Acquisition):} Operating initially at a cruising altitude of over $1000$~m, the UAV actively searches for the moving target along the black dashed trajectory. Once the target is firmly locked by the PT camera, the system smoothly switches to the tracking loop.
    \item \textbf{Phase 2 (Cooperative Loitering):} As observed in the XY trajectory, the UAV initiates a curved entry maneuver to establish a precise circular loitering pattern with the prescribed radius $R = 500$~m centered near the moving target. Concurrently, the XZ profile reveals that the UAV is commanded to descend from its initial altitude to the designated $500$~m flight ceiling. Maintaining this specific horizontal offset $R$ serves a critical geometric purpose: it effectively avoids the classical IBVS singularity issue discussed in \eqref{eq:sigular} that occurs during direct overhead flight, while the IBVS actively steers the gimbal to counteract self-occlusion risks induced by the severe roll angles during this loitering ring.
    \item \textbf{Phase 3 (Terminal Engagement):} Upon satisfying the loitering entry criteria, the control framework activates the Tracking/Terminal phase switch at approximately $X = 1200$~m. The BPNG controller takes over the autopilot commands, guiding the UAV to break away from its circular orbit. As depicted in the XZ plane, the UAV performs a highly synchronized dive, rapidly sacrificing its altitude from $500$~m down to $0$~m to intercept the maritime surface target while strictly aligning its velocity vector with the desired terminal impact angle constraint.
\end{itemize}

\begin{figure}[h]
    \centering
    \includegraphics[width=\columnwidth]{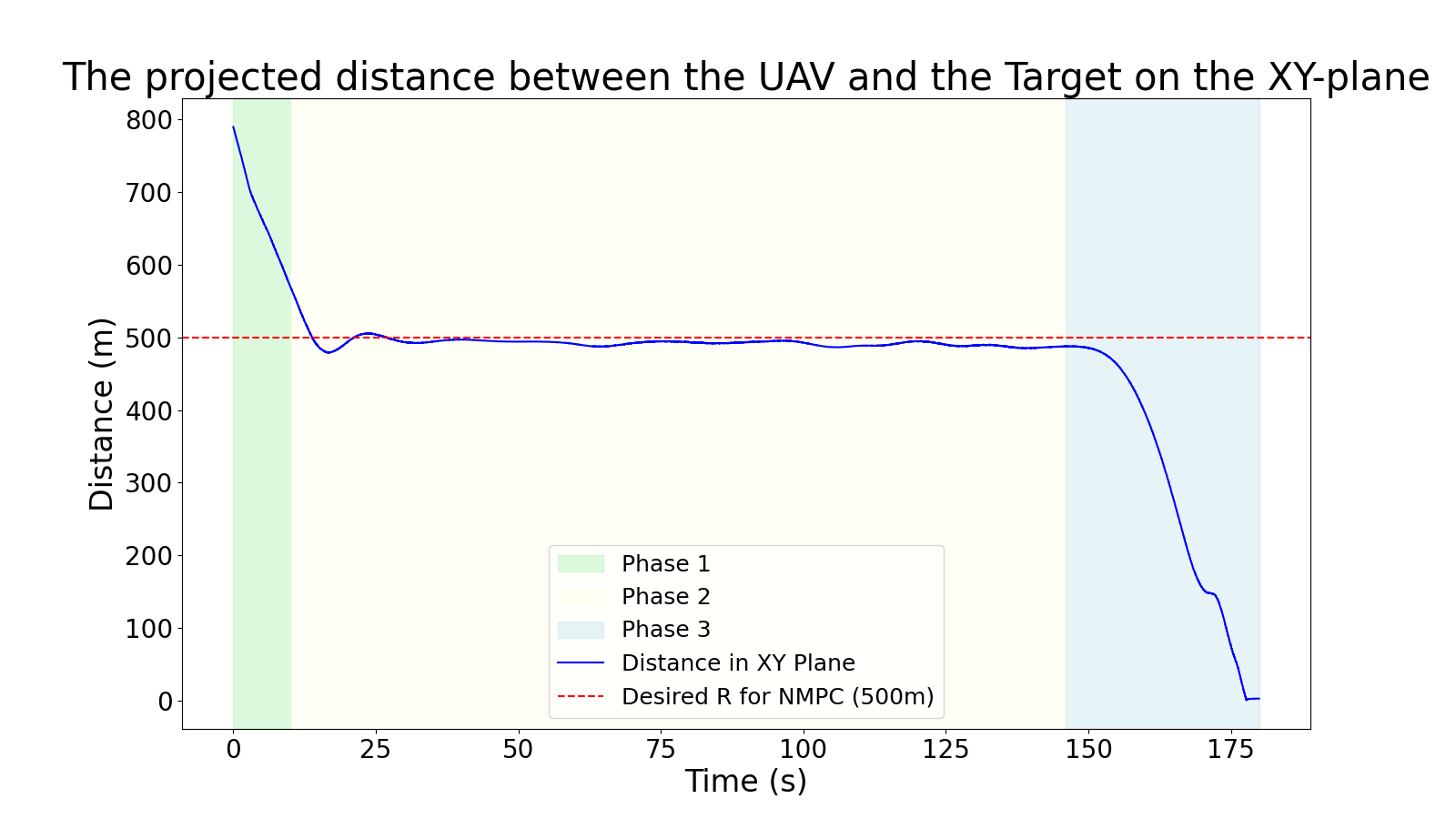}
    \caption{The distance between the target and the UAV in the XY plane.}
    \label{fig:XY distance}
\end{figure}
Figure~\ref{fig:XY distance} depicts the horizontal separation between the UAV and the target throughout the tracking phase. Beginning from a significant initial offset, the NMPC controller effectively follows the target while preserving the desired standoff distance (e.g., R=500~m). The tracking phase concludes when the UAV attains a predefined transition altitude of 500~m relative to the target—illustrated in the XZ plane in Fig.~\ref{fig:2Dplane}—at which point control authority is handed over to the BPNG controller for terminal guidance. This altitude serves as an arbitrarily chosen transition threshold rather than a tuned parameter and can be modified without requiring changes to the NMPC framework, since the controller independently steers the UAV to the prescribed terminal condition.
\begin{figure}[H]
    \centering
    \includegraphics[width=\columnwidth]{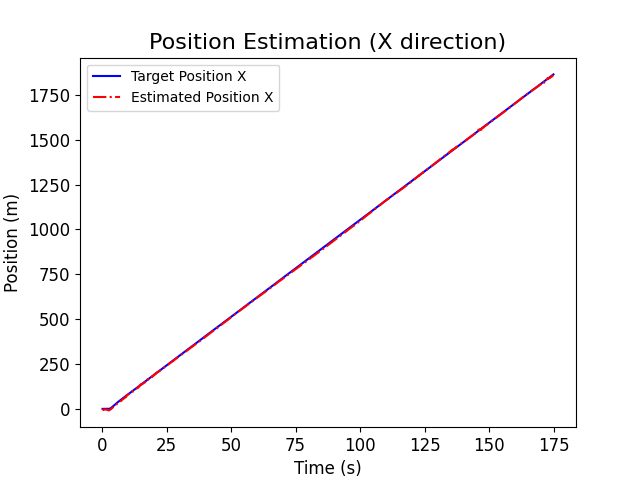}
    \caption{Comparison of the estimated target position in the $x$-direction obtained by the UKF and the ground truth.}
    \label{fig:UKF_x}
\end{figure}
\begin{figure}[H]
    \centering
    \includegraphics[width=\columnwidth]{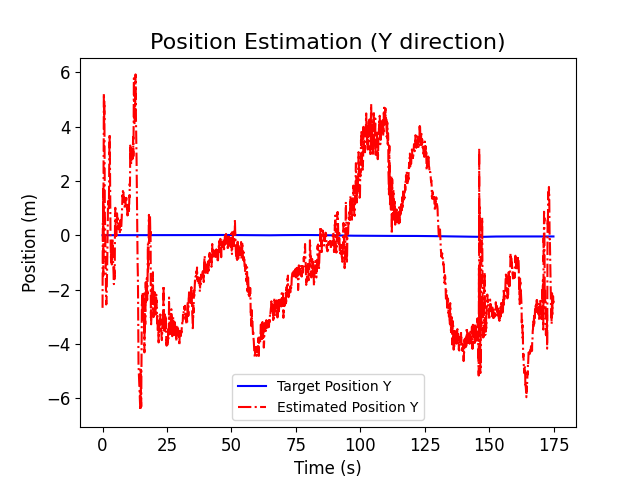}
    \caption{Comparison of the estimated target position in the $y$-direction obtained by the UKF and the ground truth.}
    \label{fig:UKF_y}
\end{figure}
\begin{figure}[H]
    \centering
    \includegraphics[width=\columnwidth]{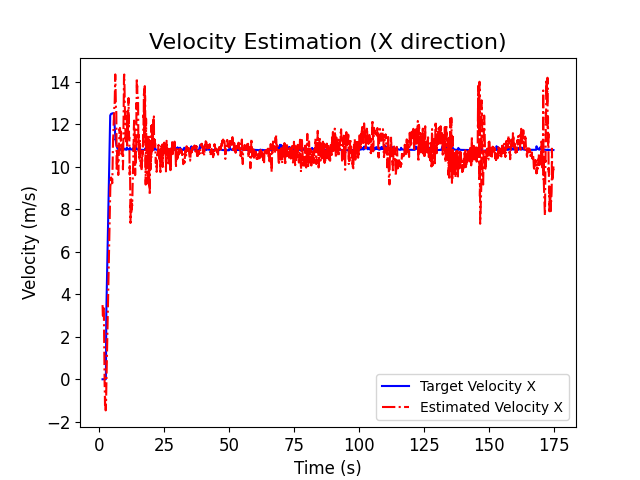}
    \caption{Comparison of the estimated target velocity in the $x$-direction obtained by the UKF and the ground truth.}
    \label{fig:UKF_x_vel}
\end{figure}
\begin{figure}[H]
    \centering
    \includegraphics[width=\columnwidth]{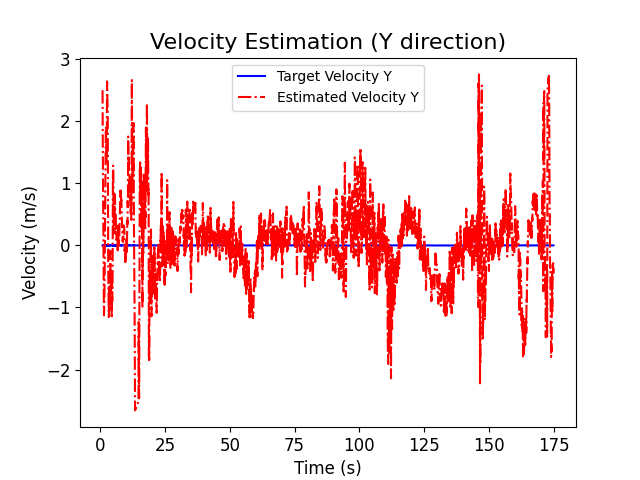}
    \caption{Comparison of the estimated target velocity in the $y$-direction obtained by the UKF and the ground truth.}
    \label{fig:UKF_y_vel}
\end{figure}
\begin{figure}[H]
    \centering
    \includegraphics[width=\columnwidth]{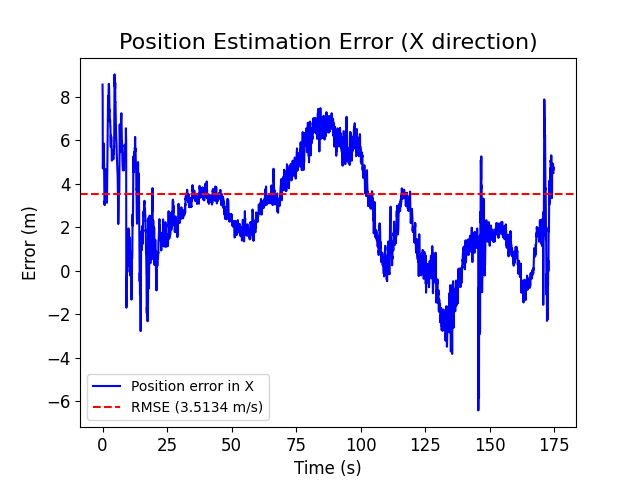}
    \caption{Position estimation error and RMSE in the $x$-direction.}
    \label{fig:RMSE_x}
\end{figure}
\begin{figure}[H]
    \centering
    \includegraphics[width=\columnwidth]{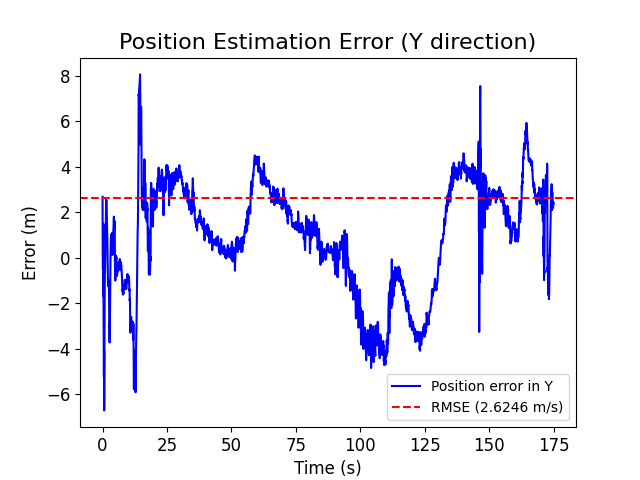}
    \caption{Position estimation error and RMSE in the $y$-direction.}
    \label{fig:RMSE_y}
\end{figure}
\begin{figure}[H]
    \centering
    \includegraphics[width=\columnwidth]{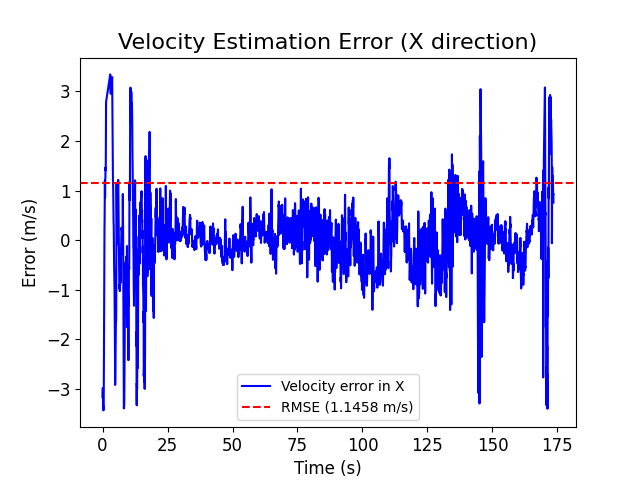}
    \caption{Velocity estimation error and RMSE in the $x$-direction.}
    \label{fig:RMSE_x_vel}
\end{figure}
\begin{figure}[H]
    \centering
    \includegraphics[width=\columnwidth]{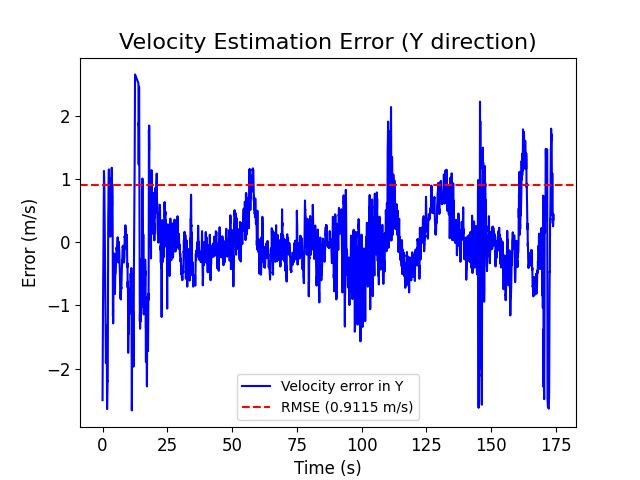}
    \caption{Velocity estimation error and RMSE in the $y$-direction.}
    \label{fig:RMSE_y_vel}
\end{figure}

The UKF-derived estimates of the target’s position and velocity are depicted in Fig.~\ref{fig:UKF_x}-\ref{fig:UKF_y_vel}, while the associated estimation errors and RMSE are shown in Fig.~\ref{fig:RMSE_x}-\ref{fig:RMSE_y_vel}. The position errors remain within 8 m, which is small compared to the surface vehicle’s dimensions (see Table \ref{tab:model_specs}) and therefore does not impair tracking performance. These results demonstrate that the UKF yields accurate and robust state estimates of the target relative to the fixed-wing UAV. After convergence is reached via persistent visual measurements, the estimated velocity aligns closely with the ground truth. In summary, the UKF successfully reconstructs the unknown target trajectory, with its effectiveness contingent on continuous visual feedback during the tracking process.
\subsection{Visual Tracking Performance}

\begin{figure}[h]
    \centering
    \subfloat{
    \includegraphics[width=\columnwidth]{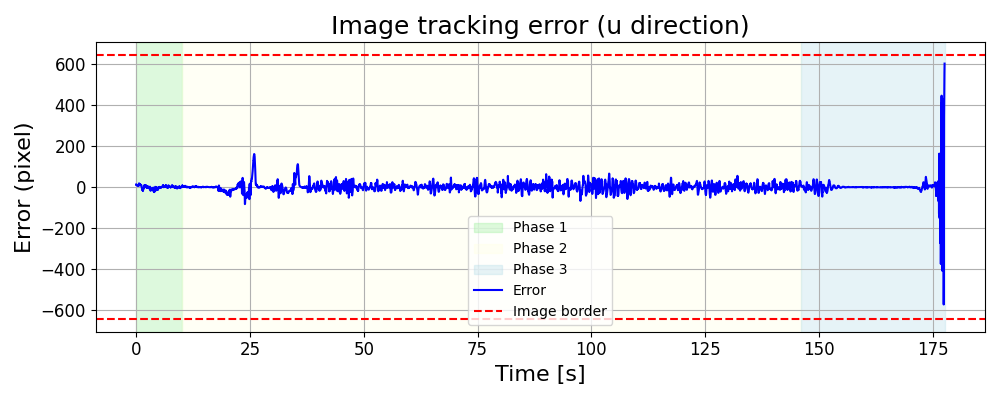}
    }\\
    \subfloat{
    \includegraphics[width=\columnwidth]{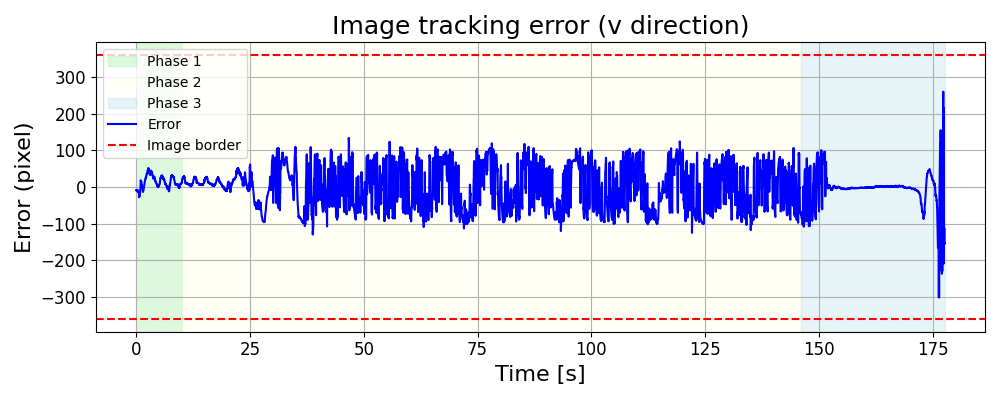}
    }
    \caption{The error between the target and the FOV center.}
    \label{fig:uv_error}
\end{figure}
\begin{figure}[h]
    \centering
    \subfloat{
    \includegraphics[width=0.9\linewidth]{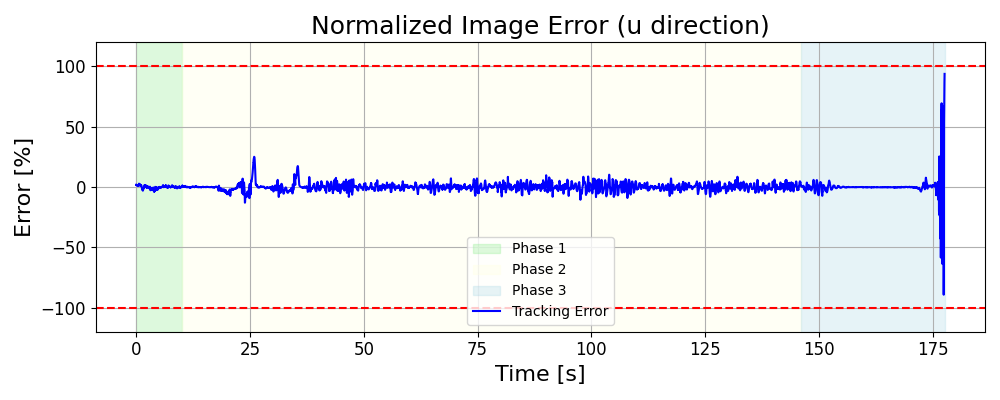}
    }\\
    \subfloat{
    \includegraphics[width=0.9\linewidth]{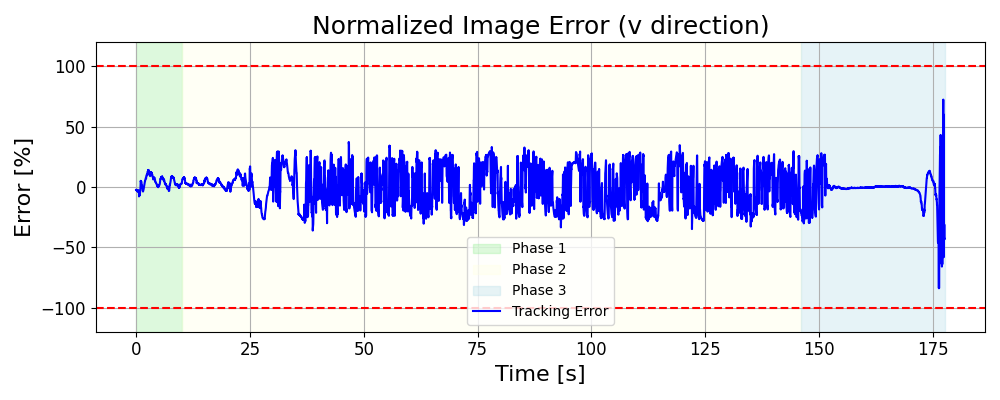}
    }
    \caption{The percentage error between the detected target and FOV center.}
    \label{fig:uv_error_pre}
\end{figure}

The tracking error metrics and the corresponding image-plane target pixel trajectories are illustrated in Fig.~\ref{fig:uv_error} and Fig.~\ref{fig:uv_error_pre}. The tracking error is quantified as the pixel displacement between the image center origin $(640, 360)$ and the target centroid bounding box detected by the YOLOv5 vision module. Given the camera's resolution of $1280 \times 720$ pixels, these figures directly evaluate the transient and steady-state authority of the IBVS loop in commanding the PT rates.

Throughout all operational phases, the IBVS controller successfully constrains the moving target well within the camera's field of view (FOV), preventing target loss. Distinct geometric characteristics can be observed between the horizontal ($u$) and vertical ($v$) error axes. The tracking error and high-frequency fluctuations in the vertical $v$-direction are noticeably larger than those in the horizontal $u$-direction. Rather than being a artifact of resolution scaling, this asymmetry is primarily attributed to the intense longitudinal coupled dynamics; the dramatic altitude variation and aggressive pitching maneuvers during the loitering transition and the subsequent terminal dive directly excite the gimbal's tilt axis. Nevertheless, the IBVS loop demonstrates rapid disturbance rejection, consistently pulling the tracking error back toward the steady-state bounds.

Particularly during the Phase~3 terminal engagement, despite the transient spikes triggered by the tracking-to-terminal controller handoff, the target pixel coordinates are kept tightly centered. These pixel measurements are continuously fed into the measurement update step of the UKF, ensuring the estimation loop receives high-rate, reliable tracking lines. These results confirm that the integrated IBVS-gimbal system achieves stable, highly robust visual tracking, providing a critical tactical cornerstone for the closed-loop guidance architecture.

\subsection{Terminal Impact Angle}
\begin{figure}[h]
    \centering
    \subfloat{
    \includegraphics[width=\columnwidth]{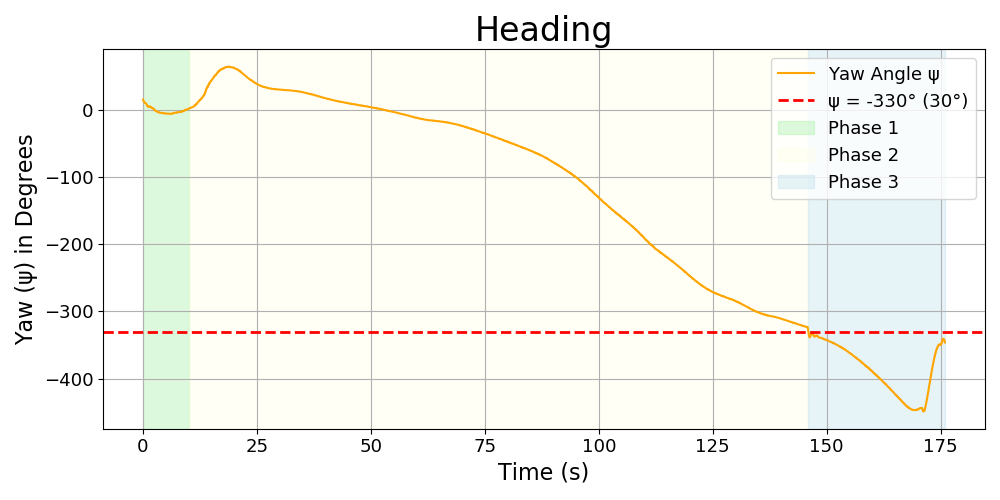}
    }\\
    \subfloat{
    \includegraphics[width=\columnwidth]{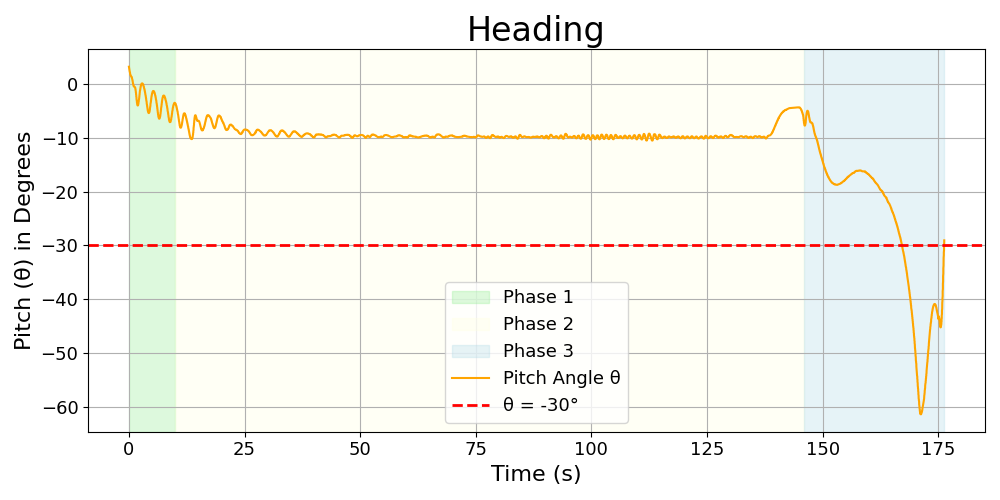}
    }
    \caption{The Euler angle of the UAV's velocity direction.}
    \label{fig:BPNG_res}
\end{figure}

In this section, the terminal engagement performance of the UAV managed by the BPNG scheme during Phase~3 is critically analyzed. Upon entering Phase~3 at $t \approx 146$~s, the guidance architecture seamlessly transitions control authority to the BPNG law to intercept the mobile ground target. Fig.~\ref{fig:BPNG_res} illustrates the temporal evolution of the UAV's velocity heading Euler angles (yaw $\psi$ and pitch $\theta$). 

Unlike stationary intercept scenarios, the moving target geometry demands highly dynamic corrections. As depicted in the profiles, the heading angles do not simply converge monotonically; instead, they exhibit substantial tactical adjustments to satisfy the strict terminal constraint vectors. In the longitudinal plane, the pitch angle undergoes a steep diving maneuver, temporarily overshooting to approximately $-60^\circ$ near $t = 171$~s to rapidly deplete altitude, before the BPNG aggressively pulls the aircraft back to precisely settle at the designated impact pitch of $\theta_f = -30^\circ$ at the final intercept instant. 

Simultaneously, the heading yaw angle corrects its loitering orientation---transitioning from the periodic wrapping profiles observed in Phase~2---and aligns tightly with the specified impact yaw of $\psi_f = 30^\circ$. These localized terminal corrections effectively demonstrate the capability of the biased proportional navigation law to decouple and satisfy both the interception requirement and the highly restricted directional impact angles simultaneously under highly dynamic simulation environments.

\subsection{Evaluation of Self-Occlusion}
To avoid self-occlusion as discussed in Section \ref{sec:self-occlusion}, the geometric condition in (\ref{eq:constraint}) constrains the UAV’s roll with respect to the target, ensuring that the PT joint remains within its operational range throughout the target tracking task. These constraints are realized in (\ref{eq:cbfright}) and (\ref{eq:cbfleft}), which convert the UAV pose restrictions into input-level constraints in the NMPC framework.

The performance of these constraints is assessed using a comparable simulation that isolates Phase~2. Two illustrative cases are examined, representing clockwise (CW) and counterclockwise (CCW) circling, where the Phase~2 NMPC enforces a circular path at a specified stand-off distance from the moving target. The direction of circling is selected based on the target’s initial lateral position with respect to the UAV.

\begin{figure}[H]
    \centering
    \includegraphics[width=\columnwidth]{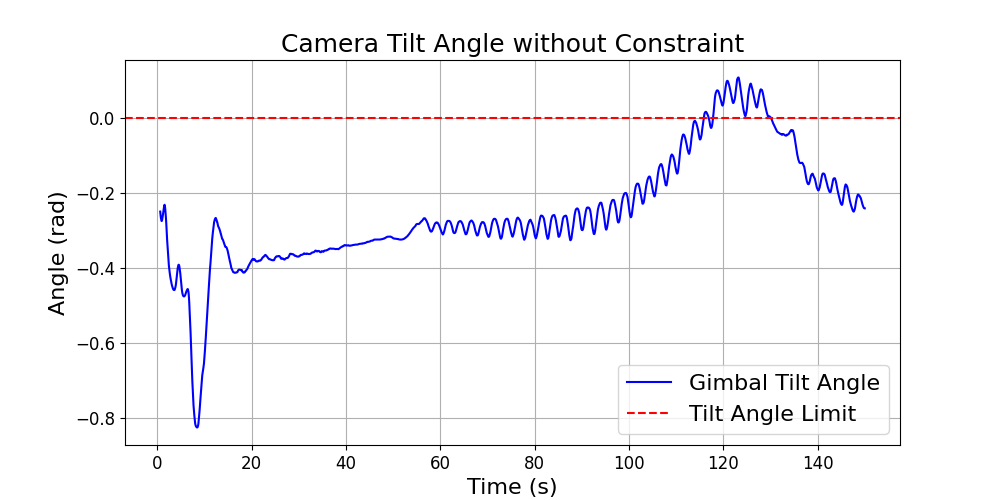}
    \caption{Tilt angle during a tracking mission carried out without applying the proposed CBF constraint.}
    \label{fig:joint}
\end{figure}

\begin{figure}[H]
    \centering
        \includegraphics[width=\columnwidth]{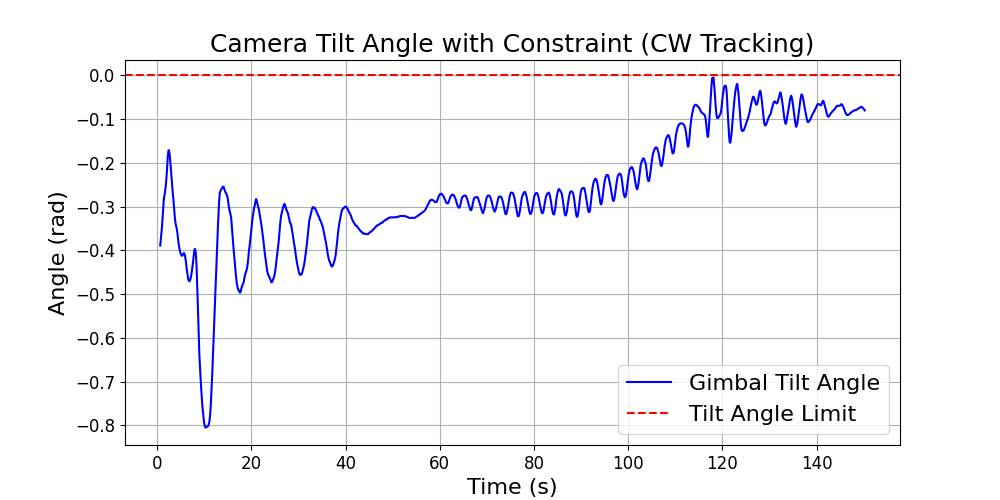}
    \caption{Tilt angle during the clockwise tracking mission subject to the proposed CBF constraint.}
    \label{fig:joint_con1}
\end{figure}

\begin{figure}[H]
    \centering
        \includegraphics[width=\columnwidth]{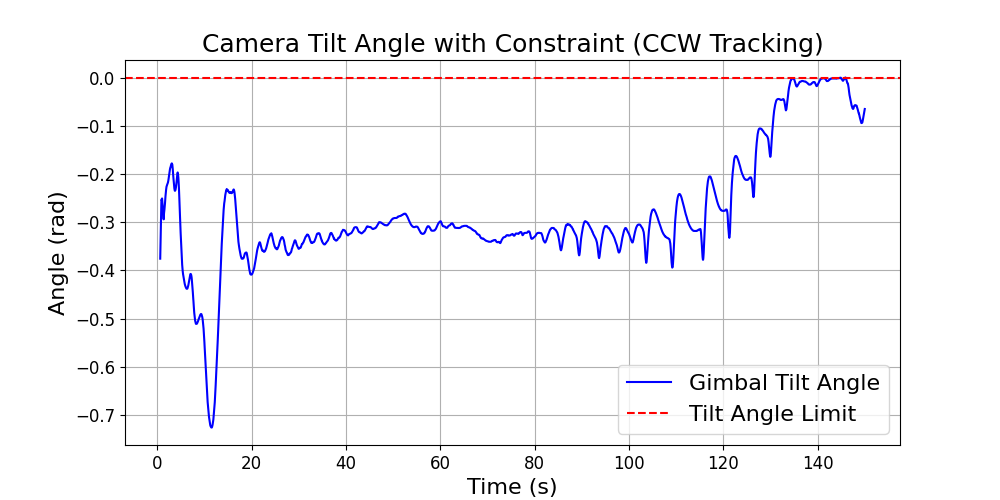}
    \caption{Tilt angle during a counterclockwise tracking mission under the proposed CBF-based constraint.}
    \label{fig:joint_con2}
\end{figure}

To evaluate the effectiveness of the CBF constraint defined in (\ref{eq:cbfright}) and (\ref{eq:cbfleft}), two scenarios are examined: operation without the CBF constraint and operation with it. The unconstrained scenario is shown in Fig.~\ref{fig:joint}, where $\theta_t$ becomes positive and the joint limit is surpassed in order to keep tracking the target, which leads to visual occlusion. In contrast, Figs.~\ref{fig:joint_con1} and~\ref{fig:joint_con2} present the  tilt angle under the proposed CBF constraint, showing that the  tilt remains within its normal operating range while keeping the target inside the camera’s FOV. This prevents UAV self-occlusion and confirms that the proposed UAV pose constraints reliably guarantee visual tracking.

\section{Conclusion}
This work presented an integrated control framework for fixed-wing UAVs, combining NMPC, IBVS, and BPNG to facilitate a seamless transition from target tracking to terminal engagement. The proposed architecture addresses the critical challenges of long-endurance tracking by fusing YOLO-based visual observations with an Unscented Kalman Filter, enabling robust state estimation of moving targets with unknown dynamics. By embedding CBFs within the NMPC, we successfully mitigated the risk of airframe-induced self-occlusion, ensuring continuous visual contact and gimbal stability throughout the loitering phase. Furthermore, the implementation of a quaternion-based BPNG law proved highly effective in guiding the UAV toward the moving target while satisfying strict impact angle constraints. High-fidelity simulation results validated the robustness of the entire framework, confirming its capability to handle realistic dynamic limits and maintain precise engagement. Future work will focus on improving resilience against temporary target loss through predictive gimbal pointing strategies and enhancing the UKF with adaptive motion models to further optimize performance in complex, dynamic maritime environments.

\bibliographystyle{ieeetr}
\bibliography{ref}

@Article{rs9111187,
AUTHOR = {Meng, Xuelian and Shang, Nan and Zhang, Xukai and Li, Chunyan and Zhao, Kaiguang and Qiu, Xiaomin and Weeks, Eddie},
TITLE = {Photogrammetric UAV Mapping of Terrain under Dense Coastal Vegetation: An Object-Oriented Classification Ensemble Algorithm for Classification and Terrain Correction},
JOURNAL = {Remote Sens.},
VOLUME = {9},
YEAR = {2017},
NUMBER = {11},
ARTICLE-NUMBER = {1187},
ISSN = {2072-4292},
DOI = {10.3390/rs9111187}
}

@article{CHAMOLA2021102324,
title = {A Comprehensive Review of Unmanned Aerial Vehicle Attacks and Neutralization Techniques},
journal = {Ad Hoc Netw.},
volume = {111},
pages = {102324},
year = {2021},
issn = {1570-8705},
author = {Vinay Chamola and Pavan Kotesh and Aayush Agarwal and  Naren and Navneet Gupta and Mohsen Guizani}
}

@ARTICLE{1310000,
  author={Sujit, P.B. and Ghose, D.},
  journal={IEEE Trans. Aerosp. Electron. Syst.}, 
  title={Search using multiple {UAVs} with flight time constraints}, 
  year={2004},
  volume={40},
  number={2},
  pages={491-509},
  doi={10.1109/TAES.2004.1310000}}

@ARTICLE{6290694,
  author={Tomic, Teodor and Schmid, Korbinian and Lutz, Philipp and Domel, Andreas and Kassecker, Michael and Mair, Elmar and Grixa, Iris Lynne and Ruess, Felix and Suppa, Michael and Burschka, Darius},
  journal={IEEE Robot. Autom. Mag.}, 
  title={Toward a Fully Autonomous {UAV}: Research Platform for Indoor and Outdoor Urban Search and Rescue}, 
  year={2012},
  volume={19},
  number={3},
  pages={46-56},
  doi={10.1109/MRA.2012.2206473}}

@INPROCEEDINGS{6761569,
  author={Ma'sum, M. Anwar and Arrofi, M. Kholid and Jati, Grafika and Arifin, Futuhal and Kurniawan, M. Nanda and Mursanto, Petrus and Jatmiko, Wisnu},
  booktitle={Proc. Int. Conf. Adv. Comput. Sci. Inf. Syst.}, 
  title={Simulation of intelligent Unmanned Aerial Vehicle ({UAV}) For military surveillance}, 
  year={2013},
  volume={},
  number={},
  pages={161-166},
  doi={10.1109/ICACSIS.2013.6761569}}

@article{oh2013rendezvous,
  title={Rendezvous and standoff target tracking guidance using differential geometry},
  author={Oh, Hyondong and Kim, Seungkeun and Shin, Hyo-Sang and White, Brian A and Tsourdos, Antonios and Rabbath, Camille Alain},
  journal={J. Intell. Robot. Syst.},
  volume={69},
  pages={389--405},
  year={2013},
  publisher={Springer}
}

@article{QUINTERO201428,
title = {Vision-based target tracking with a small {UAV}: Optimization-based control strategies},
journal = {Control Eng. Pract.},
volume = {32},
pages = {28-42},
year = {2014},
issn = {0967-0661},
author = {Steven A.P. Quintero and João P. Hespanha}
}

@article{peliti2012vision,
  title={Vision-based loitering over a target for a fixed-wing {UAV}},
  author={Peliti, Pietro and Rosa, Lorenzo and Oriolo, Giuseppe and Vendittelli, Marilena},
  journal={IFAC Proc. Vol.},
  volume={45},
  number={22},
  pages={51--57},
  year={2012},
  publisher={Elsevier}
}

@inproceedings{yang2020image,
  title={Image-based visual servo control for ground target tracking using a fixed-wing {UAV} with pan-tilt camera},
  author={Yang, Lingjie and Liu, Zhihong and Wang, Guanzheng and Wang, Xiangke},
  booktitle={Proc. Int. Conf. Unmanned Aircraft Syst.},
  pages={354--361},
  year={2020},
}

@inproceedings{mali2020model,
  title={Model predictive control for target tracking in {3D} with a downward facing camera equipped fixed wing aerial vehicle},
  author={Mali, Pravin and Singh, Arun Kumar and Krishnal, Madhav and Sujit, PB},
  booktitle={2020 IEEE CASE},
  pages={165--172},
  year={2020},
}

@inproceedings{tyagi2021nmpc,
  title={{NMPC-based UAV 3D} target tracking in the presence of obstacles and visibility constraints},
  author={Tyagi, Prakrit and Kumar, Yogesh and Sujit, PB},
  booktitle={ICUAS},
  pages={858--867},
  year={2021},
}

@inproceedings{dong2017maneuvering,
  title={Maneuvering target tracking and motion estimation using vision-aid particle filter},
  author={Dong, Fei and Zhang, Jiaqi and You, Keyou},
  booktitle={Proc. IECON Annu. Conf. IEEE Ind. Electron. Soc.},
  pages={6733--6738},
  year={2017},
}

@article{wang2014vision,
  title={{Vision-based detection and tracking of a mobile ground target using a fixed-wing UAV}},
  author={Wang, Xun and Zhu, Huayong and Zhang, Daibing and Zhou, Dianle and Wang, Xiangke},
  journal={Int. J. Adv. Robot. Syst.},
  volume={11},
  number={9},
  pages={156},
  year={2014},
  publisher={SAGE Publications Sage UK: London, England}
}

@inbook{valencia2012small,
  author    = {M. Valencia and Randal W. Beard and Timothy W. McLain},
  title     = {Small Unmanned Aircraft: Theory and Practice},
  chapter   = {Coordinate Frames}, 
  pages     = {9--15},
  publisher = {Princeton University Press},
  year      = {2012},
  isbn      = {978-06-911-4921-9}
}

@INPROCEEDINGS{8407314,
  author={Huang, Haiyang and Zhang, Lei and Fu, Chunyang and Zhou, Yonggang and Tian, Yantao},
  booktitle={Proc. Chinese Control Decis. Conf.}, 
  title={Ground moving target tracking algorithm for multi-rotor unmanned aerial vehicle}, 
  year={2018},
  volume={},
  number={},
  pages={1214-1219},
  doi={10.1109/CCDC.2018.8407314}}

@article{GomezBalderas2013,
  author    = {Gomez-Balderas, J. E. and Flores, G. and Garc{\'i}a Carrillo, L. R. and Lozano, R.},
  title     = {Tracking a Ground Moving Target with a Quadrotor Using Switching Control},
  journal   = {J. Intell. Robot. Syst.},
  year      = {2013},
  volume    = {70},
  pages     = {65--78},
  doi       = {10.1007/s10846-012-9747-9},
  publisher = {Springer},
}

@article{Ye2020,
  author    = {Ye, H. and Yang, X. and Shen, H. and Chen, Y.},
  title     = {Standoff Tracking of a Moving Target for Quadrotor Using Lyapunov Potential Function},
  journal   = {Int. J. Control Autom. Syst.},
  year      = {2020},
  volume    = {18},
  pages     = {845--855},
  doi       = {10.1007/s12555-019-0101-x},
  publisher = {Springer},
}

@phdthesis{giuffrida2018model,
  title={Model Predictive Control Techniques for fixed-wing UAV maneuvers},
  author={Giuffrida Trampetta, Federico},
  year={2018},
  school={Politecnico di Torino}
}

@article{Ulker2017,
  author    = {Ülker, Hakan and Baykara, Cemal and Özsoy, Can},
  title     = {Design of MPCs for a fixed wing {UAV}},
  journal   = {Aircr. Eng. Aerosp. Technol.},
  year      = {2017},
  volume    = {89},
  number    = {6},
  pages     = {893--901},
  doi       = {10.1108/AEAT-08-2015-0198},
  issn      = {0002-2667},
  publisher = {Emerald Publishing Limited}
}

@inproceedings{stastny2017nonlinear,
  title={Nonlinear mpc for fixed-wing {UAV} trajectory tracking: Implementation and flight experiments},
  author={Stastny, Thomas J and Dash, Adyasha and Siegwart, Roland},
  booktitle={AIAA Guid., Navig., Control Conf.},
  pages={1512},
  year={2017}
}

@INPROCEEDINGS{5611328,
  author={Low, Chang Boon},
  booktitle={IEEE Int. Conf. Control Appl.}, 
  title={A trajectory tracking control design for fixed-wing unmanned aerial vehicles}, 
  year={2010},
  volume={},
  number={},
  pages={2118-2123},
  doi={10.1109/CCA.2010.5611328}}

@inbook{stevens2015aircraft,
  title     = {Aircraft Control and Simulation: Dynamics, Controls Design, and Autonomous Systems},
  author    = {Stevens, Brian L. and Lewis, Frank L. and Johnson, Eric N.},
  year      = {2015},
  publisher = {John Wiley \& Sons},
  edition   = {3},
  chapter   = {3},
  pages     = {130--135},
  address   = {Hoboken, NJ},
  booktitle = {Aircraft Control and Simulation: Dynamics, Controls Design, and Autonomous Systems},
}

@article{yang1997unified,
  title={A unified approach to proportional navigation},
  author={Yang, Ciann-Dong and Yang, Chi-Ching},
  journal={IEEE Trans. Aerosp. Electron. Syst.},
  volume={33},
  number={2},
  pages={557--567},
  year={1997},
  publisher={IEEE}
}

@article{adler1956missile,
  title={Missile guidance by three-dimensional proportional navigation},
  author={Adler, Fred P},
  journal={J. Appl. Phys.},
  volume={27},
  number={5},
  pages={500--507},
  year={1956},
  publisher={American Institute of Physics}
}

@article{smith2008proportional,
  title={Proportional navigation with adaptive terminal guidance for aircraft rendezvous},
  author={Smith, Austin L},
  journal={J. Guid. Control Dyn.},
  volume={31},
  number={6},
  pages={1832--1836},
  year={2008}
}

@article{kim2021quaternion,
  title={Quaternion based three-dimensional impact angle control guidance law},
  author={Kim, Jeong-Hun and Park, Sang-Sup and Park, Kuk-Kwon and Ryoo, Chang-Kyung},
  journal={IEEE Trans. Aerosp. Electron. Syst.},
  volume={57},
  number={4},
  pages={2311--2323},
  year={2021},
  publisher={IEEE}
}

@inproceedings{koenig2004design,
  title={Design and use paradigms for gazebo, an open-source multi-robot simulator},
  author={Koenig, Nathan and Howard, Andrew},
  booktitle={IEEE Int. Conf. Intell. Robots Syst.},
  volume={3},
  pages={2149--2154},
  year={2004},
  organization={IEEE}
}

@ARTICLE{Junming2020,
  author={Li, Jun-Ming and Chen, Ching-Wen and Cheng, Teng-Hu},
  journal={IEEE Trans. Control Syst. Technol.}, 
  title={Motion Prediction and Robust Tracking of a Dynamic and Temporarily-Occluded Target by an Unmanned Aerial Vehicle}, 
  year={2021},
  volume={29},
  number={4},
  pages={1623-1635},
  doi={10.1109/TCST.2020.3012619}}
\end{document}